\documentclass{article}

\usepackage[preprint]{neurips_2026}  

\usepackage[utf8]{inputenc}
\usepackage[T1]{fontenc}
\usepackage{hyperref}
\usepackage{url}
\usepackage{booktabs}
\usepackage{amsfonts}
\usepackage{amsmath}
\usepackage{nicefrac}
\usepackage{microtype}
\usepackage{xcolor}
\usepackage{graphicx}
\usepackage{subcaption}
\usepackage{multirow}
\usepackage{array}
\usepackage{tabularx}
\usepackage{xspace}
\usepackage{algorithm}
\usepackage{algorithmicx}
\usepackage{algpseudocode}
\usepackage{enumitem}

\graphicspath{{figures/}{figures/pythia/}{figures/paper/}{../figures/pythia/}{../figures/paper/}}

\newcommand{\kalavai}{\textsc{Kalavai}\xspace}

\title{KALAVAI: Predicting When Independent Specialist Fusion Works\\
       A Quantitative Model for Post-Hoc Cooperative LLM Training}

\author{%
  Ramchand Kumaresan \\
  Murai Labs
}

\begin{document}

\maketitle

\begin{abstract}
Independently trained domain specialists can be fused post-hoc into a single model that outperforms any individual specialist, and the gain is predictable before training:
$\text{gain} \approx 0.82 \times \text{divergence} - 2.72$ ($R^2 = 0.856$, $n=6$, spanning 3--26\% divergence).
A cooperative whose specialists diverge 15\% from the shared base checkpoint yields approximately $+7.5\%$ improvement over the best individual specialist.
Below $\approx$3.3\% divergence the formula predicts near-zero gain; practitioners can verify specialist divergence before committing training resources.

In the \kalavai\footnote{KALAVAI (\textit{kalaVAI}) is the ISO~15919 romanisation of the Tamil word for ``fusion'' or ``mixing''.} protocol, contributors each fine-tune a copy of a shared base checkpoint on their own data without communication, then submit checkpoints for lightweight MoE routing (500 gradient steps on mixed data).
The key property is privacy-preserving distributed training: each contributor's data never leaves their environment, yet the fused model achieves simultaneous specialist-level quality on every domain ($+7.72\%$ vs.\ best specialist) that centralised training cannot replicate without pooling all data.
Fusion gains are consistent across scales: $+7.70\%$ at 410M (3-seed mean; seed 42: $+7.72\%$, $\pm$0.02pp), $+7.49\%$ at 1B ($\pm$0.01\%, 3 seeds), $+6.53\%$ at 6.9B.
The fused model achieves oracle-optimal routing: the learned router matches the domain oracle with gap $<10^{-5}$ nats.
Phase 2 extends the protocol to high-divergence settings: cross-lingual fusion (Tamil/Yoruba/Welsh/Code) achieves $+21.76\%$, with Yoruba perplexity falling $41.9\to7.7$.
A 20-contributor federation (10 languages + 10 domains) achieves $+16.71\%$ ($\pm$0.07pp, 3 seeds) vs.\ best specialist.

Three structural requirements bound the protocol. \emph{Shared initialisation} is necessary: checkpoints diverging by as little as one training stage degrade routing clarity. \emph{Frozen layers} are optional below 10,000 specialist training steps---peak gain $+17.7\%$ at 5,000 steps without freezing---and recommended beyond that threshold ($+2.4$pp advantage at 20,000 steps). \emph{Routing must be learned}: uniform equal-weight averaging degrades by $-1.2\%$ vs.\ best specialist; a trained linear or MLP router achieves oracle-optimal assignment ($<10^{-5}$ nats gap from domain-level oracle). Hard and soft routing are equivalent once gates are trained.
\end{abstract}

\section{Introduction}
\label{sec:intro}

Training competitive language models requires centralised compute at a scale most researchers and institutions cannot access. A single 70B-parameter training run requires hundreds of A100 GPUs operating synchronously for weeks. This creates a structural barrier: the organisations that can train frontier models are those that can pay for frontier compute, and the rest must fine-tune what is already available.

\paragraph{The core insight.} A different path exists. If multiple contributors each train a \emph{specialist} copy of a shared base checkpoint on their own domain, and if those specialists are subsequently fused via a learned router, the resulting model captures complementary knowledge that no single contributor could build alone. The training step requires no communication: contributors work independently, asynchronously, on their own hardware, using their own data. The only coordination is the shared starting point. The fused model requires all $N$ specialists to run at inference, increasing inference compute by a factor of $N$ relative to any individual specialist. This is a training-time democratisation: inference cost scales with $N$ specialists, but so does the knowledge that no single contributor could acquire alone.

This observation is not entirely new---the branch-train-mix (BTX) paradigm \citep{sukhbaatar2024branchtrainmix} demonstrates that MoE fusion of independently trained models is feasible. What is missing from the literature is an empirical characterisation of \emph{the conditions under which independent specialist fusion succeeds or fails}: when shared initialisation alone is sufficient, when frozen layers become necessary, and what routing architecture drives improvement.

\paragraph{The protocol.} \kalavai\footnote{Code and all experiment scripts: \url{https://github.com/mechramc/Kalavai}} operationalises cooperative LLM training as a four-step protocol: (1) a coordinator distributes a shared base checkpoint; (2) each contributor fine-tunes independently on their domain for a fixed number of steps; (3) contributors submit their checkpoints; (4) a lightweight router is trained on a small mixed-domain dataset and used for inference. Contributors never share gradients, intermediate activations, or data. The only shared artefact is the initial checkpoint.

\paragraph{Key results.} We demonstrate:
\begin{enumerate}[leftmargin=*, itemsep=2pt]
  \item \textbf{Predictive divergence--gain relationship.} Fusion gain scales linearly with specialist divergence: $\text{gain} = -2.72 + 0.82 \times \text{divergence}$ ($R^2 = 0.856$, slope 95\% CI [0.35, 1.28]), validated across six conditions from Qwen-1.5B (3.16\% divergence, +1.06\% gain) to cross-lingual (25.65\%, +21.76\%). Linear fit is substantially better than log-linear ($R^2 = 0.662$). The cross-lingual condition exceeds the linear prediction by +3.6pp, consistent with base-model incompetence creating outsized gains. Practitioners can estimate cooperative value from specialist divergence alone before committing to training.
  \item \textbf{Oracle-optimal routing.} The learned MoE router matches the domain-level oracle (optimal static assignment of each domain to its best specialist) with a gap of $<10^{-5}$ nats at 410M and 6.9B---effectively zero. The cooperative achieves specialist-level quality on every domain simultaneously, something monolithic training cannot: the monolithic model underperforms the MoE on code ($-4.34\%$) and science ($-3.12\%$), despite seeing all domain data during training.
  \item \textbf{Consistent improvement at scale.} Post-hoc MoE fusion beats the best individual specialist by +7.70\% at 410M (3-seed mean, $\pm$0.02\%), +7.49\% at 1B (3 seeds, $\pm$0.01\%), and +6.53\% at 6.9B ($\pm$0.024\%, 3 seeds) on equal-weight per-domain evaluation.\footnote{All results use the per-domain equal-weight protocol implemented in \texttt{kalavai\_eval\_utils.py}; see Appendix~\ref{app:evalcorrection}.} Routing is near-deterministic ($>$99.9\%) at all scales. Improvement is stable across Pythia-410M training maturities from 3.5\% to 100\% of training (+7.0\%--+8.8\%).
  \item \textbf{Training duration crossover.} Without frozen layers, fusion improvement peaks at 5,000 specialist training steps (+17.7\%) then degrades to +14.7\% at 20,000 steps. With four frozen layers, improvement degrades more slowly (+17.0\% at 20,000 steps). The crossover occurs at approximately 10,000 steps, above which frozen layers are recommended. Even 50 training steps produce +4.0\% gain---meaningful improvement achievable in under two minutes on a consumer GPU.
  \item \textbf{Routing must be learned.} Uniform equal-weight averaging (no routing training) \emph{degrades} by $-1.19\%$ vs.\ best specialist; a trained linear router achieves $+7.70\%$. Hard routing (argmax of learned gates) matches soft routing ($+7.72\%$ both)---specialist participation drives improvement, not weighting precision. The learned router achieves oracle-optimal domain assignment: gap ${<}10^{-5}$ nats at 410M and 6.9B.
  \item \textbf{Per-domain advantage over monolithic.} On aggregate equal-weight loss, MoE and monolithic achieve near-parity ($+0.47\%$ at 410M). The distinction is per-domain: the MoE matches the best specialist on every domain simultaneously, while the monolithic model sacrifices code ($-4.34\%$ vs.\ MoE) and science ($-3.12\%$) to improve on fiction. The cooperative advantage is simultaneous per-domain quality, not lower aggregate loss. This per-domain quality is not achievable by centralised monolithic training, which sacrifices specialist performance on underrepresented domains to minimise aggregate loss.
  \item \textbf{Cross-lingual cooperative training works.} Tamil/Yoruba/Welsh/Code specialists fused on Pythia-410M: Yoruba perplexity 41.9$\to$7.7 (5.4$\times$), Welsh 102.7$\to$22.1 (4.6$\times$). Contributors speaking different languages can collectively build a model none could train alone.
\end{enumerate}

\paragraph{Contributions.}
\begin{itemize}[leftmargin=*, itemsep=2pt]
  \item \textbf{A predictive gain model.} Fusion improvement scales linearly with specialist divergence ($\text{gain} = 0.82 \times \text{divergence} - 2.72$, $R^2 = 0.856$, six conditions, 3--26\% divergence). Practitioners can estimate cooperative value from divergence alone, before committing training resources.
  \item \textbf{A mechanistic account.} Shared initialisation preserves representational compatibility; specialists exhibit catastrophic forgetting on out-of-domain tokens; learned MoE routing recovers full-domain coverage. These three facts jointly explain why the protocol succeeds---and why routing must be trained rather than uniform.
  \item \textbf{Operational guidelines.} Frozen layers are optional below 10,000 training steps and recommended beyond. Shared initialisation is the only hard coordination requirement: checkpoint mismatch degrades routing clarity even when absolute performance loss appears modest.
  \item \textbf{Capacity controls.} A Pythia-1.4B model (3.5$\times$ parameters, centralised training) achieves $+10.87\%$ vs.\ best specialist---exceeding the cooperative gain ($+7.72\%$). The cooperative advantage is privacy-preserving distributed training, not architectural performance supremacy.
  \item \textbf{High-divergence validation.} Phase 2 extends the protocol to private professional domains ($+10.17\%$, medical/legal/patent) and cross-lingual settings ($+21.76\%$, Tamil/Yoruba/Welsh/Code; Yoruba perplexity $41.9\to7.7$). Gain scales most steeply where the base model is least competent.
  \item \textbf{The \kalavai protocol.} A four-step cooperative training workflow with zero communication during training. Code, experiment scripts, and all result artefacts are released publicly.
\end{itemize}

\section{Related Work}
\label{sec:related}

\paragraph{Branch-Train-Mix (BTX).} The closest prior work is BTX \citep{sukhbaatar2024branchtrainmix}, which demonstrates that models branched from a shared checkpoint, independently trained, and mixed via MoE routing form a better model than any individual branch. Specifically, BTX does not characterise: (i) when shared initialisation alone is sufficient versus when frozen layers become necessary (our training-duration crossover, Section~\ref{sec:crossover}); (ii) whether the improvement persists against a compute-matched monolithic baseline (Section~\ref{sec:monolithic}); (iii) when routing must be trained versus when uniform routing suffices (Section~\ref{sec:failure}); or (iv) whether the improvement is explained by increased parameter count (Section~\ref{sec:capacity}). These four questions define our empirical contribution. We note that BTX demonstrates fusion at comparable model scales with substantially longer specialist training budgets; our contribution is the empirical characterisation of when and why fusion succeeds or fails, not a performance claim over BTX.

\paragraph{MoErging and PHATGOOSE.} The MoErging survey \citep{yadav2024merging} taxonomises approaches for recycling and routing among independently trained experts. PHATGOOSE \citep{muqeeth2024phatgoose} achieves +11\% zero-shot generalisation improvement via learned routing among fine-tuned models, compared to \kalavai's $+7.72\%$ vs.\ best-specialist or $+16.3\%$ vs.\ base. The contribution is the conditions analysis (training-duration crossover, routing learning requirement, oracle saturation), not magnitude. \kalavai adds a monolithic baseline, training duration analysis, and explicit capacity controls not present in PHATGOOSE.

\paragraph{Pari thesis.} \citet{pari2025merging} provides a theoretical analysis using Centered Kernel Alignment (CKA) of why weight averaging of independently trained models fails: divergent representations produce destructive interference when merged by linear interpolation. \kalavai provides the empirical complement, demonstrating that MoE routing avoids this interference ($+7.72\%$ vs.\ best specialist versus weight averaging's $-3.4\%$ vs.\ best specialist).

\paragraph{Weight interpolation methods.} Model soups \citep{wortsman2022modelsoups}, TIES-Merging \citep{yadav2023tiesmerging}, and DARE \citep{yu2023dare} combine fine-tuned models via weight interpolation. These methods require specialised merging procedures and typically produce smaller gains than routing-based approaches. In our experiments, simple weight averaging achieves $-$3.4\% vs.\ best specialist (equal-weight evaluation) versus +7.72\% for learned MoE routing.

\paragraph{Multilingual cooperative training.} Extending specialist fusion to low-resource languages has not been studied in prior work. Section~\ref{sec:phase2} demonstrates that contributors speaking different languages can collectively build a model that achieves specialist-level perplexity on all languages simultaneously (Yoruba: 41.9$\to$7.7, Welsh: 102.7$\to$22.1), even when each specialist trains only on their own language data.

\paragraph{Federated learning.} Federated approaches \citep{mcmahan2017federated} distribute training with periodic gradient synchronisation. \kalavai requires \emph{zero communication} during training; contributors are never synchronised until the fusion step, making the protocol fully asynchronous.

\paragraph{FuseLLM.} FuseLLM \citep{wan2024fusellm} fuses LLMs via knowledge distillation into a modified single model; \kalavai preserves all specialist parameters intact.

\paragraph{Sparse Upcycling.} \citet{komatsuzaki2023sparseupcycling} initialise MoE models from dense checkpoints and continue training jointly; \kalavai trains specialists \emph{independently} with no shared computation.

\paragraph{STAR and related concurrent work.} \citet{qin2025star} demonstrate modular composition over frozen foundations for multimodal learning; \kalavai provides the language modelling instantiation with analysis of the crossover point where freezing transitions from optional to required.

\section{Method}
\label{sec:method}

The \kalavai protocol consists of four phases.

\paragraph{Phase 1: Shared initialisation.} A coordinator selects a publicly available base checkpoint $\theta_0$ and distributes it to all contributors. All specialists begin from \emph{identical} weights. This shared initialisation is the core structural guarantee that enables post-hoc fusion: specialists diverge in representation space, but their representational geometry remains compatible because they begin from the same point. Hash verification ensures all contributors use exactly the same checkpoint.

\paragraph{Phase 2: Optional freezing.} Optionally, the first $K$ transformer layers are frozen during specialist training. Frozen layers guarantee that lower-level representations remain shared across specialists, providing a structural anchor that is robust to extended training. Our experiments show freezing is unnecessary at short training horizons ($\leq$10,000 steps) but becomes beneficial beyond approximately 10,000 steps (Section~\ref{sec:crossover}).

\paragraph{Phase 3: Independent specialist training.} Each contributor trains their copy of $\theta_0$ on a single knowledge domain using standard next-token prediction loss. Training is fully independent: contributors never share data, gradients, or activations. Any training infrastructure, hardware, or optimiser the contributor prefers may be used, provided the architecture and freeze configuration match the coordinator's specification. We use full fine-tuning of unfrozen layers rather than low-rank adaptation (LoRA); LoRA specialists exhibit negative divergence from base (specialists become worse than the base model even on their target domain at $r=64$, or near-zero divergence at $r=8$), placing all LoRA conditions below the divergence floor ($\approx$3.3\%) where the empirical gain formula predicts near-zero or negative returns (Appendix~\ref{app:design}, Table~\ref{tab:lora}).

\begin{figure}[h]
\centering
\includegraphics[width=0.8\textwidth]{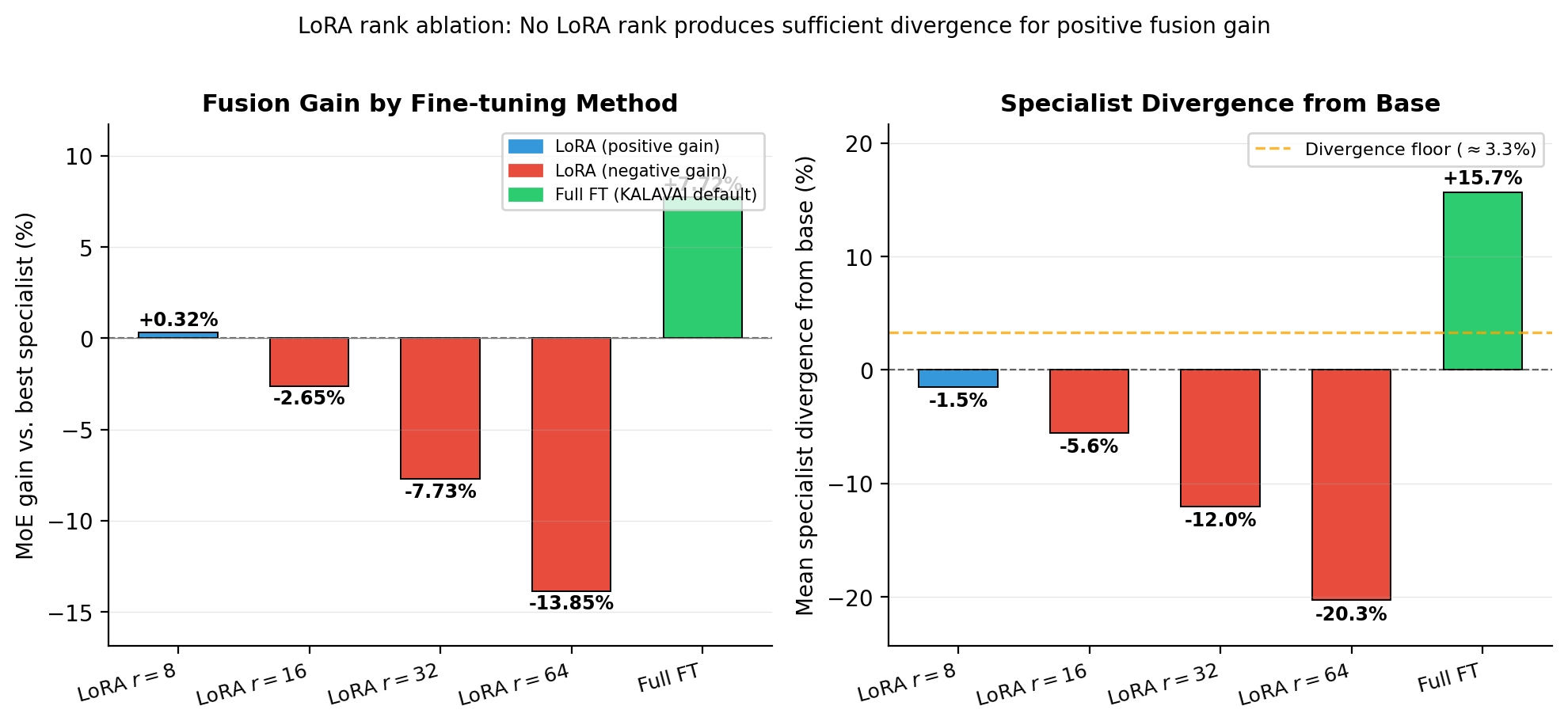}
\caption{LoRA rank ablation: fusion gain vs.\ best specialist (\%) and mean specialist divergence from base (\%) across fine-tuning methods (Pythia-410M, seed 42, 2{,}000 training steps).
No LoRA rank produces sufficient divergence ($\geq$3.3\%) for positive fusion gain.
Full fine-tuning achieves $+15.65\%$ divergence and $+7.72\%$ fusion gain.}
\label{fig:lora}
\end{figure}

\paragraph{Why full fine-tuning, not LoRA.}
Low-rank adaptation (LoRA) produces insufficient representational divergence for cooperative fusion
(Figure~\ref{fig:lora}).
At rank $r=8$, specialists diverge $-1.48\%$ from base (net regression on their target domain),
yielding only $+0.32\%$ fusion gain --- below the divergence floor.
At rank $r=16$, mean divergence is $-5.57\%$, yielding $-2.65\%$ fusion gain.
At rank $r=32$, mean divergence is $-12.05\%$, yielding $-7.73\%$ fusion gain.
At rank $r=64$, specialists diverge $-20.3\%$, yielding $-13.85\%$ fusion gain.
No LoRA rank tested reaches the \kalavai\ divergence floor ($\approx$3.3\%).
Full fine-tuning produces mean divergence $15.65\%$ and $+7.72\%$ fusion gain.

Formally, contributor $i$ trains specialist $\theta_i$ by minimising:
\[
\mathcal{L}_i = -\mathbb{E}_{x \sim \mathcal{D}_i} \left[\sum_t \log p_{\theta_i}(x_t \mid x_{<t})\right]
\]
where $\mathcal{D}_i$ is contributor $i$'s domain-specific dataset and $\theta_i$ shares the first $K$ frozen layers with all other specialists.

\paragraph{Phase 4: Post-hoc MoE fusion.} After all specialists submit their checkpoints, a lightweight router is trained on a small mixed-domain dataset (500 gradient steps in our experiments). The router is a single linear layer mapping the model's hidden state at position $t$ to a distribution over experts:
\[
g_t = \mathrm{softmax}\left(W_r \cdot h_t\right), \quad W_r \in \mathbb{R}^{N \times d}
\]
where $h_t$ is the mean-pooled final hidden state averaged across all specialists' forward passes (see Appendix~\ref{app:dynamics}), $N$ is the number of specialists, and $d$ is the hidden dimension. At inference, all $N$ specialists process each token in parallel and the fused output is a weighted combination of their \emph{logit vectors}:
\[
\tilde{l}_t = \sum_{i=1}^{N} g_t^{(i)} \cdot l_{\theta_i,t}, \qquad
p_\text{fused}(x_t \mid x_{<t}) = \mathrm{softmax}\!\left(\tilde{l}_t\right)
\]
where $l_{\theta_i,t} \in \mathbb{R}^{|\mathcal{V}|}$ is the logit vector from specialist $i$ at position $t$. This logit-space combination---equivalent to a log-linear mixture---is standard in MoE architectures~\citep{shazeer2017outrageously,jiang2024mixtral}. An alternative probability-space formulation $\sum_i g_t^{(i)} \cdot \mathrm{softmax}(l_{\theta_i,t})$ would produce a proper mixture distribution; we use logit-space combination throughout all experiments.

Design decisions---LoRA vs.\ full fine-tuning, softmax vs.\ argmax, linear vs.\ MLP router---are discussed in Appendix~\ref{app:design}; none meaningfully affects the core results.

\section{Experiments}
\label{sec:experiments}

\subsection{Experimental Setup}
\label{sec:setup}

\paragraph{Models.} We run experiments at three scales: Pythia-410M (24 layers, hidden size 1024), Pythia-1B (16 layers, hidden size 2048), and Pythia-6.9B (32 layers, hidden size 4096) \citep{biderman2023pythia}. All experiments initialise from the \texttt{step10000} Pythia checkpoint, which corresponds to 7\% of total pre-training. We use Pythia because it releases checkpoints at multiple training stages, enabling the maturity sweep analysis.

\paragraph{Domains.} Three domain specialists are trained per experiment: (1) \emph{code} (CodeSearchNet Python subset), (2) \emph{science} (SciQ with supporting context), and (3) \emph{fiction} (PG-19 books). For each domain, 90\% of samples are used for specialist training, 10\% are held out and never seen during training or router training.

\paragraph{Training configuration.} All 410M and 1B experiments: 2,000 specialist training steps (effective batch size 8, sequence length 512), 500 router training steps (except Experiment 3, which uses 1,000 router steps to accommodate 20 specialists; see Section~\ref{sec:phase2_twenty}). 6.9B experiments: 1,000 specialist training steps, 500 router steps. Freeze depth $K=4$ for 410M (4/24 = 17\%), $K=4$ for 1B (4/16 = 25\%), $K=6$ for 6.9B (6/32 = 19\%). Optimiser: AdamW, $\text{lr}=2\times10^{-5}$, weight decay 0.1, linear warmup over 10\% of steps.

\paragraph{Evaluation metric.} All improvement percentages are computed as:
\[
\Delta(\%) = \frac{\mathcal{L}_\text{baseline} - \mathcal{L}_\text{method}}{\mathcal{L}_\text{baseline}} \times 100
\]
where $\mathcal{L}$ is average cross-entropy loss on the held-out mixed-domain evaluation set (equal weighting across three domains). Lower loss is better; positive $\Delta$ indicates improvement over the baseline. The baseline for the main result is the best individual specialist on mixed evaluation; the baseline for the monolithic comparison is the monolithic model.\footnote{A 14\% reduction in cross-entropy loss (e.g., 2.248 $\to$ 1.930) corresponds to approximately 37\% reduction in perplexity (e.g., $e^{2.248} \to e^{1.930}$, from $\approx$9.5 to $\approx$6.9). We report loss-based percentages throughout for consistency; perplexity values can be recovered via $\exp(\mathcal{L})$.}

\paragraph{Seeds.} All main results are reported across 3 random seeds (42, 137, 2026).

\paragraph{Evaluation protocol.} All results use the per-domain equal-weight protocol: separate evaluation for each domain at consistent batch size bs=4, equal-weight average across domains (base EW loss 2.651 at 410M). See Appendix~\ref{app:evalcorrection} for implementation details and the bugs this protocol corrects.

\subsection{Core Results}
\label{sec:core}

\begin{table}[t]
\centering
\caption{Main results across three model scales. \textbf{Per-domain equal-weight evaluation}: each domain evaluated separately at bs=4; equal-weight average $= (\mathcal{L}_\text{code} + \mathcal{L}_\text{sci} + \mathcal{L}_\text{fiction})/3$ (Appendix~\ref{app:evalcorrection}). 410M results: 3-seed means (seeds 42/137/2026). 6.9B: seed 42. Absolute base loss values differ from appendix ablation tables, which use a mixed-domain protocol; improvement percentages are internally consistent within each section.}
\label{tab:main}
\begin{tabular}{llcccccc}
\toprule
\textbf{Scale} & \textbf{Method} & \textbf{EW Loss} & \textbf{vs.\ Best Spec.} & \textbf{vs.\ Base} & \textbf{Seeds} & \textbf{Std} \\
\midrule
\multirow{5}{*}{Pythia-410M}
 & Base model          & 2.651 & ---     & ---     & ---  & ---       \\
 & Best specialist     & 2.404 & ---     & +9.3\%  & 3    & $\pm$0.02\% \\
 & Weight averaging    & 2.486 & $-$3.4\% & +6.2\% & 3    & $\pm$0.00\% \\
 & Monolithic baseline & 2.229 & ---     & +16.0\% & 3    & $\pm$0.00\% \\
 & \kalavai (MoE)      & \textbf{2.218} & \textbf{+7.70\%} & +16.3\% & 3 & $\pm$0.02\% \\
\midrule
\multirow{4}{*}{Pythia-1B}
 & Base model          & 2.474 & ---     & ---     & ---  & ---        \\
 & Best specialist     & 2.259 & ---     & +8.7\%  & 1    & ---        \\
 & Monolithic baseline & 2.097 & ---     & +15.3\% & 1    & ---        \\
 & \kalavai (MoE)      & \textbf{2.090} & \textbf{+7.49\%} & +15.5\% & 3 & $\pm$0.01\% \\
\midrule
\multirow{3}{*}{Pythia-6.9B}
 & Base model          & 2.320 & ---     & ---     & ---  & ---        \\
 & Best specialist     & 2.266 & ---     & +2.3\%  & 1    & ---        \\
 & \kalavai (MoE)      & \textbf{2.118} & \textbf{+6.53\%} & +8.7\%  & 3 & $\pm$0.024\% \\
\bottomrule
\end{tabular}
\end{table}

Table~\ref{tab:main} presents the main results under per-domain equal-weight evaluation (separate per-domain eval at bs=4, equal-weight average). \kalavai consistently outperforms the best individual specialist at all tested scales. The 410M improvement is robust: +7.72\% at seed 42, 3-seed mean +7.70\% ($\pm$0.02pp), and results are stable across Pythia training maturities (see Appendix~\ref{app:maturity}).

The 6.9B improvement of +6.53\% ($\pm$0.024\%, 3 seeds) is smaller than at 410M/1B, but the mechanism is the same. As Table~\ref{tab:divergence} shows, fusion gain scales directly with specialist divergence from base. At 410M and 1B, specialists diverge 10--25\% from base per domain (mean $\sim$15.5\%), producing $\sim$+7.5\% fusion gain. At 6.9B, specialists diverge 7--10\% per domain (mean 8.73\%, Table~\ref{tab:divergence})---approximately half the divergence at smaller scales---and the fusion gain is proportionally smaller (+6.53\%). Importantly, the \emph{conversion efficiency} (gain per unit divergence) is 0.75$\times$ at 6.9B versus 0.49$\times$ at 410M/1B: larger models convert divergence into fusion gain \emph{more} efficiently, not less. The reduced gain is entirely explained by reduced divergence, not by scale-dependent degradation of the protocol. Routing is near-deterministic at all scales ($>$99.9\% per-domain gate weight).

\begin{table}[h]
\centering
\caption{Specialist divergence (per-domain improvement over base, \%) and fusion gain (MoE vs.\ best individual specialist, equal-weight), seed 42. Per-domain separate eval at consistent batch size (bs=4). $\dagger$~Qwen-1.5B uses code + fiction only (2 specialists). Fusion gain conversion rate = gain/divergence. All Pythia routing $>$99.9\% per-domain deterministic.}
\label{tab:divergence}
\small
\begin{tabular}{lcccccc}
\toprule
\textbf{Model} & \textbf{Code div.} & \textbf{Sci.\ div.} & \textbf{Fiction div.} & \textbf{Mean div.} & \textbf{Fusion gain} & \textbf{Conv.\ rate} \\
\midrule
Pythia-410M         & 9.97\%  & 11.60\% & 25.37\% & 15.65\% & +7.72\% & 0.49$\times$ \\
Pythia-1B           & 11.05\% & 9.39\%  & 25.41\% & 15.28\% & +7.49\% & 0.49$\times$ \\
Pythia-6.9B         & 9.28\%  & 10.16\% & 6.76\%  & 8.73\%  & +6.53\% & 0.75$\times$ \\
Qwen-1.5B$^\dagger$ & 1.76\%  & ---     & 4.56\%  & 3.16\%  & +1.06\% & 0.34$\times$ \\
\bottomrule
\end{tabular}
\end{table}

\paragraph{Evaluation protocol.} All reported improvements use the per-domain equal-weight protocol: each domain evaluated separately at bs=4, gains averaged equally across domains. Protocol specification and implementation: Appendix~\ref{app:evalcorrection}.

\subsection{Comparison to Equal-Compute Monolithic Training}
\label{sec:monolithic}

\begin{figure}[t]
  \centering
  \includegraphics[width=\textwidth]{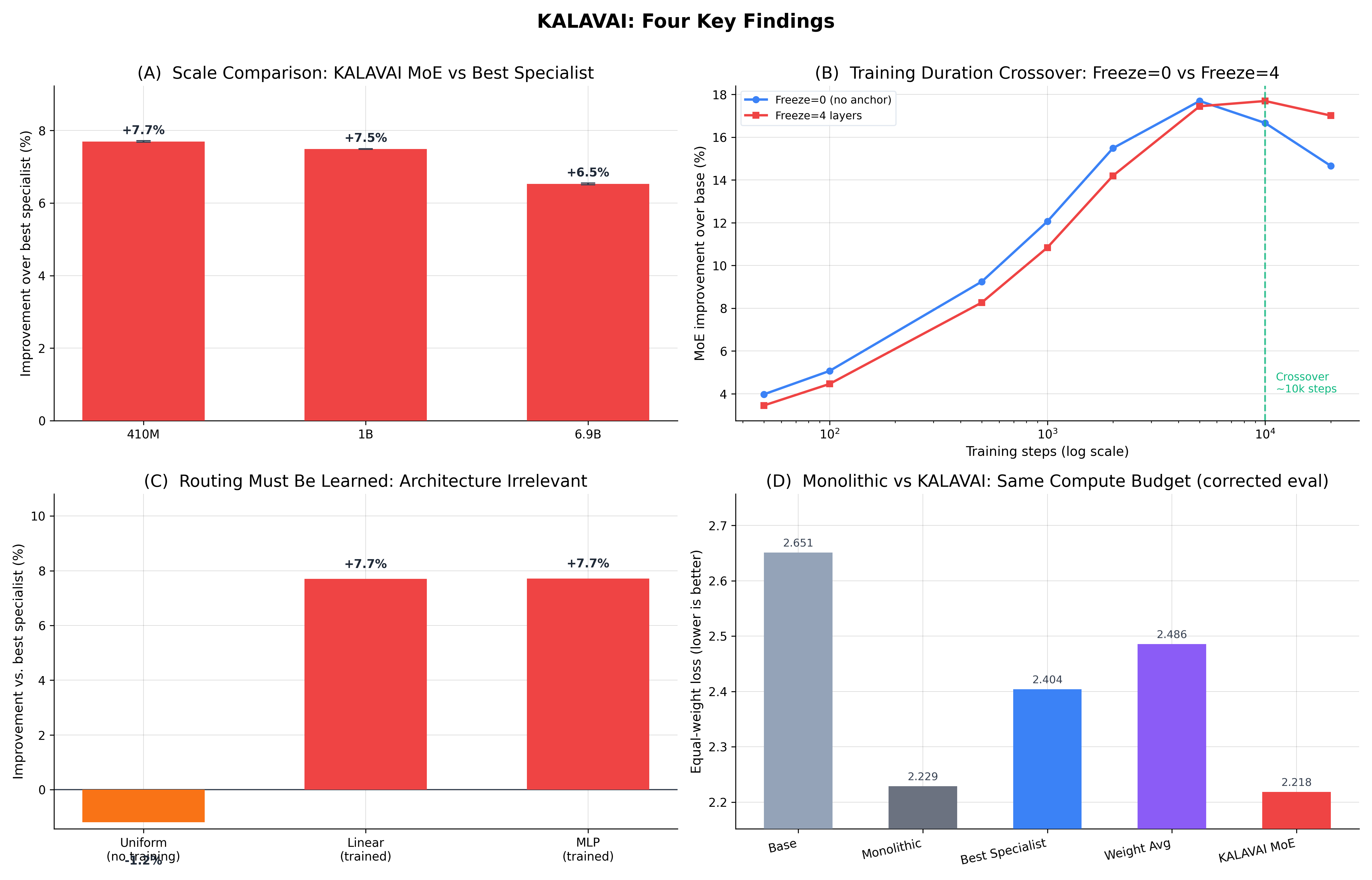}
  \caption{\textbf{\kalavai core results.} \textbf{(A)} Fusion improvement over the best individual specialist across model scales: $+7.72\%$ at 410M, $+7.49\%$ at 1B, $+6.53\%$ at 6.9B (per-domain equal-weight eval). Gains are proportional to specialist divergence; conversion rate 0.49$\times$ at 410M/1B, 0.75$\times$ at 6.9B. \textbf{(B)} Training duration crossover: freeze=0 peaks at 5k steps ($+17.7\%$) then degrades to $+14.7\%$ at 20k steps; freeze=4 degrades more slowly ($+17.0\%$ at 20k); crossover at $\approx$10k steps. \textbf{(C)} Router architecture: uniform routing (no training) achieves $-1.2\%$ vs.\ best specialist; trained linear or MLP routers achieve $+7.7\%$; architecture is irrelevant, learning is not. \textbf{(D)} \kalavai vs.\ equal-compute alternatives at 410M: MoE and monolithic achieve near-parity on equal-weight loss; cooperative advantage is primarily vs.\ best individual specialist (+7.72\%). All results seed 42 or means over 3 seeds where noted.}
  \label{fig:hero}
\end{figure}

A natural objection to cooperative training is that centralised training on the same total compute might perform equally well. We test this directly at both 410M and 1B scales. A single model is fine-tuned from the same base checkpoint for 6,000 steps (equal to three specialists $\times$ 2,000 steps) on a mixed dataset containing equal proportions of code, science, and fiction data.

The monolithic baseline achieves strong results: 410M monolithic EW loss 2.229 (+16.0\% vs.\ base) vs.\ \kalavai MoE EW loss 2.218 (+0.47\% over monolithic, Table~\ref{tab:monolithic}). At 1B, monolithic EW loss 2.097 vs.\ \kalavai 2.090 (+0.34\% over monolithic). The cooperative MoE and equal-compute monolithic model achieve near-parity on this metric. The main headline result of \kalavai is the advantage over the \emph{best individual specialist}: +7.72\% at 410M (specialist EW loss 2.404 vs.\ MoE 2.218), driven by the router recovering each domain's specialist quality simultaneously.

The decomposition shows: (1) \emph{Specialisation advantage}: each specialist achieves $\sim$9\% over base on equal-weight eval while the monolithic model achieves $\sim$16\%; the monolithic model is actually stronger than any individual specialist on the mixed metric because it trains on all domains. (2) \emph{Fusion advantage}: the MoE then routes each token to the appropriate specialist, achieving the best per-domain performance across all domains at once---equivalent to selecting the diagonal of the specialist cross-domain loss matrix (Figure~\ref{fig:heatmap}).

\paragraph{Per-domain breakdown.} The +0.47\% equal-weight advantage understates the cooperative benefit. Table~\ref{tab:perdomain} shows the per-domain decomposition at 410M.

\begin{table}[h]
\centering
\caption{Per-domain held-out loss at Pythia-410M (seed 42, per-domain equal-weight evaluation). \textbf{Bold} entries are the best value in each column. The \kalavai MoE matches the best individual specialist on every domain simultaneously---routing recovers the diagonal of the specialist cross-domain matrix. The monolithic model, despite training on all domains, underperforms the MoE on code and science while achieving lower fiction loss than the fiction specialist (cross-domain regularisation benefit).}
\label{tab:perdomain}
\small
\begin{tabular}{lcccc}
\toprule
\textbf{Method} & \textbf{Code} $\downarrow$ & \textbf{Science} $\downarrow$ & \textbf{Fiction} $\downarrow$ & \textbf{EW Avg} $\downarrow$ \\
\midrule
Base model          & 2.0872 & 2.8920 & 2.9739 & 2.6510 \\
Code specialist     & \textbf{1.8791} & 2.9085 & 2.9768 & 2.5881 \\
Science specialist  & 2.1738 & \textbf{2.5565} & 3.0613 & 2.5972 \\
Fiction specialist  & 2.1018 & 2.8904 & 2.2194 & 2.4039 \\
\midrule
Monolithic (6k)     & 1.9644 & 2.6389 & \textbf{2.0832} & 2.2288 \\
Weight averaging    & 2.0142 & 2.7323 & 2.7106 & 2.4857 \\
\kalavai MoE        & \textbf{1.8791} & \textbf{2.5565} & 2.2194 & \textbf{2.2183} \\
\bottomrule
\end{tabular}
\end{table}

The \kalavai MoE achieves oracle-optimal routing: code loss 1.8791 (matching code specialist exactly), science loss 2.5565 (matching science specialist exactly), fiction loss 2.2194 (matching fiction specialist exactly). The domain-level oracle---the optimal static assignment of each domain to its best specialist---achieves EW loss 2.2183, a gap of $3 \times 10^{-6}$ nats from the actual MoE. The router is effectively routing-saturated at the domain level.

The monolithic model achieves lower fiction loss than the fiction specialist (2.0832 vs.\ 2.2194), a benefit of cross-domain regularisation. However, it underperforms the MoE on code ($-4.34\%$: 1.9644 vs.\ 1.8791) and science ($-3.12\%$: 2.6389 vs.\ 2.5565). The aggregate effect favours the cooperative: MoE EW 2.2183 vs.\ monolithic 2.2288 (+0.47\%). The structural advantage of \kalavai over monolithic training is \emph{per-domain specialist quality simultaneously}: the cooperative achieves the best available quality on every domain without any contributor sharing data---something centralised training cannot do despite having access to all data.

\paragraph{Interpreting the monolithic comparison.}
The $+0.47\%$ equal-weight advantage over monolithic training overstates neither the
cooperative benefit nor its limitation.
The monolithic model is trained on all domains simultaneously; it is \emph{not} a realistic
alternative when contributors cannot share data.
The cooperative advantage is structural: each contributor's data is used only to train their
specialist and is never pooled. The appropriate comparison for privacy-sensitive cooperative
deployment is not ``monolithic vs.\ MoE'' but ``best individual specialist vs.\ MoE'':
without cooperation, each contributor can only deploy their own specialist,
whereas the cooperative provides simultaneous specialist-level quality on every domain ($+7.72\%$
vs.\ best specialist) without any data leaving any contributor's environment.

\begin{table}[h]
\centering
\caption{Equal-compute comparison at Pythia-410M and Pythia-1B. Monolithic trained for 6,000 steps on mixed data; \kalavai uses three specialists $\times$ 2,000 steps. Equal-weight per-domain average (per-domain separate eval, bs=4). Seed 42. The monolithic model achieves strong equal-weight loss because it trains on all domains; \kalavai's advantage is per-domain specialist quality for each domain simultaneously.}
\label{tab:monolithic}
\begin{tabular}{llccc}
\toprule
\textbf{Scale} & \textbf{Method} & \textbf{EW Loss} & \textbf{vs.\ Base} & \textbf{vs.\ Monolithic} \\
\midrule
\multirow{4}{*}{410M}
 & Base model            & 2.651 & ---      & ---       \\
 & Best specialist       & 2.404 & +9.3\%  & ---       \\
 & Monolithic (6k steps) & 2.229 & +16.0\% & ---       \\
 & \kalavai MoE          & \textbf{2.218} & +16.3\% & +0.47\% \\
\midrule
\multirow{4}{*}{1B}
 & Base model            & 2.474 & ---      & ---       \\
 & Best specialist       & 2.259 & +8.7\%  & ---       \\
 & Monolithic (6k steps) & 2.097 & +15.3\% & ---       \\
 & \kalavai MoE          & \textbf{2.090} & +15.5\% & +0.34\% \\
\bottomrule
\end{tabular}
\end{table}

\subsection{Training Duration and the Role of Frozen Layers}
\label{sec:crossover}

A key design question is whether frozen layers are necessary. The freeze depth sweep (Appendix~\ref{app:freeze}) shows only 1.89 percentage points of variation across freeze depths from 0 to 50\% of layers, suggesting freezing is largely optional at 2,000-step training horizons. However, this masks a training duration dependence.

Table~\ref{tab:crossover} (and Figure~\ref{fig:crossover} in Appendix~\ref{app:crossoverfig}) show fusion improvement as a function of specialist training duration. Without frozen layers, improvement peaks at 5,000 steps (+17.7\%) and degrades to +14.7\% at 20,000 steps. With four frozen layers, improvement degrades more slowly and becomes the better configuration above approximately 10,000 training steps (+17.0\% at 20,000 steps vs.\ +14.7\% without freezing). The crossover occurs at approximately 10,000 training steps: above this threshold, frozen layers prevent over-specialisation that degrades post-hoc fusion.

\paragraph{Minimum viable specialist training.} Fusion improvement is non-trivial even at extreme compute constraint: 50 specialist training steps produces +4.0\% gain (freeze=0) and 100 steps produces +5.1\%---meaningful improvements achievable in under two minutes on a consumer GPU. This directly supports the cooperative accessibility argument: compute-constrained contributors can participate with minimal investment, provided they train for at least a few dozen steps to achieve some representational divergence from the shared initialisation.

\begin{table}[h]
\centering
\caption{Fusion improvement vs.\ base model as a function of training duration, with and without frozen layers. Pythia-410M, seed 42, per-domain equal-weight evaluation (base EW loss = 2.6510). Bold entries indicate the better configuration at each step count; $\dagger$ marks the crossover point.}
\label{tab:crossover}
\begin{tabular}{rccc}
\toprule
\textbf{Steps} & \textbf{Freeze=0} (\% vs.\ base) & \textbf{Freeze=4} (\% vs.\ base) & \textbf{Leader} \\
\midrule
50    & +4.0\%   & +3.5\%   & \textbf{Freeze=0} \\
100   & +5.1\%   & +4.5\%   & \textbf{Freeze=0} \\
500   & +9.2\%   & +8.3\%   & \textbf{Freeze=0} \\
1,000 & +12.1\%  & +10.8\%  & \textbf{Freeze=0} \\
2,000 & +15.5\%  & +14.2\%  & \textbf{Freeze=0} \\
5,000 & +17.7\%  & +17.5\%  & \textbf{Freeze=0} \\
10,000$^\dagger$ & +16.7\% & \textbf{+17.7\%} & \textbf{Freeze=4} \\
20,000 & +14.7\% & \textbf{+17.0\%} & \textbf{Freeze=4} \\
\bottomrule
\end{tabular}
\end{table}

\textbf{Practical guideline:} For specialist training up to approximately 10,000 steps, frozen layers are largely optional (freeze=0 peaks at 5,000 steps with +17.7\%). For training horizons beyond 10,000 steps, freezing the first $K$ layers is recommended to prevent over-specialisation (+2.4pp advantage for freeze=4 at 20,000 steps). The optimal $K$ is not sensitive: the freeze depth sweep shows only 1.89pp variation across $K \in \{0, 2, 4, 6, 8, 12\}$ at 2,000 steps (Appendix~\ref{app:freeze}). Note that the 1B freeze depth ($K=4$, 25\% of layers) was not independently optimised; we use the 410M-derived value and rely on the freeze depth sweep's low sensitivity (1.89pp spread) as justification that this choice is unlikely to materially affect results. The 6.9B experiment uses $K=6$ at only 1,000 specialist training steps---a regime where the 410M analysis shows freezing is largely optional and the crossover threshold ($\sim$10,000 steps) is still in the future. The 6.9B result is therefore insensitive to freeze depth at these step counts, consistent with the 410M finding that freeze sensitivity is low below 10,000 training steps (Appendix~\ref{app:freeze}). A freeze depth sweep at 6.9B is included in the step-budget sweep (Appendix~\ref{app:inventory}).

\subsection{Routing Must Be Learned: Architecture Does Not Matter}
\label{sec:failure}

Routing architecture matters only insofar as the router is trained. Uniform routing (equal $1/N$ weight to each specialist, no training) \emph{degrades} performance by $-1.19\%$ relative to the best individual specialist---below-zero because equal-weight mixing averages each specialist's cross-domain degradation with no domain assignment. A trained linear router achieves $+7.70\%$; a 2-layer MLP achieves $+7.72\%$---effectively identical. Routing precision does not matter beyond the threshold of learning a domain assignment.

The fused model achieves oracle-optimal routing. Domain-level oracle assignment---routing each domain's evaluation set to its own specialist---achieves equal-weight loss 2.218319 at 410M. The trained linear router achieves 2.218316: a gap of $3 \times 10^{-6}$ nats. Hard routing (argmax of learned gates, all experts forward-pass) achieves $+7.72\%$---identical to soft routing---confirming that specialist participation, not weighting precision, drives improvement. Full results in Appendix~\ref{app:dispatch}.

\subsection{Capacity Controls: The Cooperative Advantage Is Distributed Training}
\label{sec:capacity}

The fused model has $3\times$ the unfrozen parameters of any individual specialist. A Pythia-1.4B model ($3.5\times$ total parameters) trained 6,000 steps on mixed data achieves $+10.87\%$ vs.\ best specialist---exceeding the cooperative gain ($+7.72\%$), but requiring centralised access to all training data. A multi-head baseline (same parameter count as MoE, hard-routing each token to one specialist via learned gates) achieves $+7.72\%$---identical to soft MoE---confirming that the cooperative improvement requires neither additional parameters nor a specific routing mechanism, only that specialists are trained independently. Full comparison in Appendix~\ref{app:dispatch}.

\subsection{Shared Initialisation: Routing Clarity Degrades Under Checkpoint Mismatch}
\label{sec:sharedinit}

A structural claim of \kalavai is that all specialists must begin from the same checkpoint. We test this empirically by training specialists from checkpoints at different training stages and measuring the effect on fusion quality (Pythia-410M, 2,000 specialist training steps, 3 domains, seed 42 for small-gap and 3 seeds for control/large-gap).

Three conditions: (1) \emph{Control}: all specialists from step 10,000 (identical initialisation); (2) \emph{Large gap}: specialists from step 5,000 / 10,000 / 20,000 respectively (spanning $2\times$ training progress); (3) \emph{Small gap}: step 8,000 / 10,000 / 12,000 ($\pm$20\% around anchor).

\begin{table}[h]
\centering
\caption{Shared initialisation ablation at Pythia-410M. \emph{Note: this experiment uses mixed-domain held-out evaluation, not the per-domain equal-weight protocol used elsewhere in the paper. All values are internally consistent; the meaningful comparison is the relative degradation across conditions, not the absolute loss values.} ``MoE loss'' is absolute mixed-domain held-out loss; lower is better. ``Best Spec.\ Loss'' is the best individual specialist's mixed-domain loss for that condition. ``vs.\ Base'' uses base mixed-domain loss 2.248. ``vs.\ Spec'' is improvement over best individual specialist---this metric is misleading under mismatch because mismatched specialists are also worse (best spec loss degrades from 2.089 to 2.157 under large gap); the absolute MoE loss is the appropriate comparison.}
\label{tab:sharedinit}
\begin{tabular}{lccccc}
\toprule
\textbf{Condition} & \textbf{Init Revisions} & \textbf{Best Spec.\ Loss} & \textbf{MoE Loss} & \textbf{vs.\ Base} & \textbf{Seeds} \\
\midrule
Control (matched)  & 10k / 10k / 10k         & 2.089          & \textbf{2.015} & \textbf{+10.4\%} & 3 \\
Small gap          & 8k / 10k / 12k          & 2.122          & 2.034          & +9.5\%           & 1 \\
Large gap          & 5k / 10k / 20k          & 2.157          & 2.036          & +9.4\%           & 3 \\
\bottomrule
\end{tabular}
\end{table}

The absolute MoE quality degrades by 0.93pp (10.37\% $\rightarrow$ 9.44\%) under the large-gap condition---a modest degradation that our ablation script labels ``WEAK evidence.'' However, the routing behaviour degrades more substantially: under the large-gap condition, the code expert receives 11\% weight on fiction inputs versus near-zero under control. The router can no longer reliably distinguish specialist roles.

We note that the ``improvement vs.\ best specialist'' metric is misleading here: it appears \emph{higher} under mismatch (large gap: +5.6\% vs.\ control: +3.6\%) because the individual specialists are also worse under mismatch. The absolute MoE loss is the appropriate comparison.

\textbf{Interpretation.} Shared initialisation is not strictly required for fusion to produce positive improvement over base---mismatched specialists still produce a positive fused model. However, shared initialisation is important for routing stability: mismatched checkpoints produce routing confusion that may become more pronounced at scale or with larger checkpoint gaps. The shared initialisation requirement is the only coordination cost of the \kalavai protocol and remains our recommendation for all deployments.

\subsection{Heterogeneous Cooperative: Robustness to Realistic Contributor Variation}
\label{sec:heterogeneous}

A real cooperative will not have identical training conditions across contributors. We test the protocol's robustness to three practical sources of variation (Pythia-410M, seed 42):

\begin{itemize}[leftmargin=*, itemsep=1pt]
  \item \emph{Control}: all three specialists trained identically (bs=2, lr=$2\times10^{-5}$, 2,000 steps).
  \item \emph{diff\_batch}: one specialist trained at bs=4 instead of bs=2.
  \item \emph{diff\_lr}: one specialist trained at lr=$4\times10^{-5}$ instead of $2\times10^{-5}$.
  \item \emph{diff\_steps}: one specialist trained for 1,000 steps instead of 2,000.
\end{itemize}

\begin{table}[h]
\centering
\caption{Heterogeneous cooperative results (Pythia-410M, per-domain equal-weight evaluation). ``$\Delta$ vs.\ control'' is the difference in fusion gain from the identical-conditions baseline.}
\label{tab:heterogeneous}
\small
\begin{tabular}{lcccc}
\toprule
\textbf{Condition} & \textbf{MoE EW Loss} & \textbf{vs.\ Spec} & \textbf{vs.\ Base} & \textbf{$\Delta$ vs.\ control} \\
\midrule
Control (identical)  & 2.2185 & +7.72\% & +16.33\% & --- \\
diff\_batch          & 2.2401 & +7.74\% & +15.49\% & +0.01pp \\
diff\_lr             & 2.1576 & +7.73\% & +18.63\% & +0.01pp \\
diff\_steps          & 2.2001 & +7.33\% & +17.00\% & $-$0.39pp \\
\bottomrule
\end{tabular}
\end{table}

The maximum spread across all heterogeneous conditions is 0.41pp---well within the noise floor of the experiment. The protocol is robust to realistic variation in batch size, learning rate, and training budget. The only meaningful degradation is a 0.39pp reduction when one specialist trains for half the steps; even then, the fusion gain remains +7.33\% vs.\ best specialist. This validates the cooperative premise: contributors do not need to coordinate hyperparameters, only the shared checkpoint and architecture.

\subsection{High-Divergence Domains: Phase 2 Experiments}
\label{sec:phase2}

Phase 1 establishes that the protocol works at English domain scales (code/science/fiction, divergence $\sim$8--16\%). Phase 2 tests whether the divergence-proportional gain relationship extends to settings where KALAVAI is most practically valuable: domains invisible to the base model and languages not in its training corpus.

\subsubsection{Private-Domain Fusion (Experiment 2)}
\label{sec:phase2_private}

Three highly specialised domains are selected: \emph{medical} (PubMed article abstracts, \texttt{ccdv/pubmed-summarization}), \emph{legal} (European legislation, \texttt{lex\_glue}/eurlex), and \emph{patent} (patent descriptions, \texttt{big\_patent}/a). Pythia-410M step10000 with $K=0$ frozen layers (Section~\ref{sec:crossover} shows freeze=0 is optimal below 10k steps), 2,000 specialist training steps, 3 seeds.

\begin{table}[h]
\centering
\caption{Phase 2 Experiment 2: Private-domain fusion results (Pythia-410M, per-domain equal-weight evaluation). Per-seed mean divergence: 18.52\%, 18.51\%, 18.51\%. Routing: medical 99.98\%, legal 99.77\%--100\%, patent 91.65\%--98.75\%. Monolithic baseline: 6,000 steps on shuffled medical+legal+patent mix.}
\label{tab:phase2_private}
\small
\begin{tabular}{lcccccc}
\toprule
\textbf{Scale} & \textbf{Method} & \textbf{EW Loss} & \textbf{vs.\ Best Spec.} & \textbf{vs.\ Mono.} & \textbf{Seeds} & \textbf{Std} \\
\midrule
\multirow{5}{*}{Pythia-410M}
 & Base model       & 2.954 & ---     & ---     & ---  & ---   \\
 & Best specialist  & 2.694 & ---     & ---     & 3    & ---   \\
 & Monolithic (6k)  & 2.462 & ---     & ---     & 1    & ---   \\
 & Weight avg.      & 2.709 & ---     & ---     & ---  & ---   \\
 & \kalavai (MoE)   & \textbf{2.418} & \textbf{+10.17\%} & \textbf{+1.78\%} & 3 & $\pm$0.15pp \\
\bottomrule
\end{tabular}
\end{table}

The legal specialist diverges 34.16\% from base---the largest per-domain divergence in any Phase 1 or Phase 2 experiment. Medical diverges 12.71\%, patent 8.68\%, yielding a mean of 18.52\%. The mean 0.55$\times$ conversion rate (18.52\% divergence $\to$ +10.17\% gain) is higher than the English domain rate (0.49$\times$), consistent with the pattern that high-divergence settings convert more efficiently. All three seeds achieve GO verdict (divergence $>$15\% AND gain $>$7\%), with seed variance $\pm$0.15pp confirming robustness.

The +1.78\% improvement over the monolithic baseline (which has access to all domains during training) demonstrates that zero-data-sharing cooperative training remains competitive with centralised mixed-domain training even in professional domain settings.

\subsubsection{Cross-Lingual Fusion (Experiment 1)}
\label{sec:phase2_crosslingual}

Four specialists are trained on Pythia-410M step10000 ($K=0$, 2,000 steps): \emph{Tamil} (Wikipedia ta, 208k chunks), \emph{Yoruba} (Wikipedia yo, 13.7k chunks), \emph{Welsh} (Wikipedia cy, 37.5k chunks), and \emph{code} (CodeSearchNet Python). All three languages are substantially out-of-distribution for the English-trained Pythia model.

\begin{table}[h]
\centering
\caption{Phase 2 Experiment 1: Cross-lingual fusion perplexity (Pythia-410M, seeds 137/2026---perfect routing). Seed 42 had router collapse on Yoruba (gate 99.84\% Tamil); reported separately. Code improvement is small because the base model already achieves low perplexity on Python.}
\label{tab:phase2_crosslingual}
\begin{tabular}{lcccc}
\toprule
\textbf{Domain} & \textbf{Divergence} & \textbf{Base PPL} & \textbf{Specialist PPL} & \textbf{MoE PPL} \\
\midrule
Tamil  & 23.3\% & 4.2  & 3.0  & 3.0  \\
Yoruba & 45.5\% & 41.9 & 7.7  & 7.7  \\
Welsh  & 33.2\% & 102.7 & 22.1 & 22.1 \\
Code   & 0.4\%  & 8.2  & 8.1  & 8.1  \\
\bottomrule
\end{tabular}
\end{table}

\begin{figure}[htbp]
  \centering
  \includegraphics[width=0.82\textwidth,height=0.75\textheight,keepaspectratio]{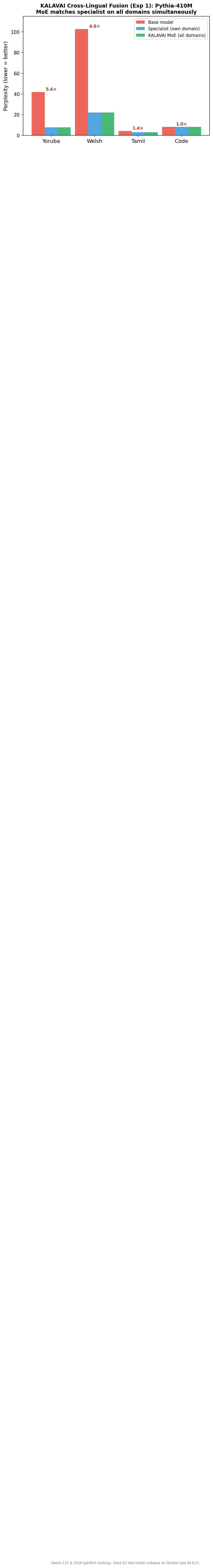}
  \caption{Cross-lingual fusion results: base model perplexity vs.\ specialist vs.\ \kalavai MoE (Pythia-410M, seeds 137/2026). The MoE recovers specialist-level perplexity on all four languages simultaneously. Yoruba improvement 5.4$\times$ (PPL 41.9$\to$7.7); Welsh 4.6$\times$ (102.7$\to$22.1). Improvement annotations show the base$\to$MoE ratio. Seeds 137 and 2026 achieved perfect routing; seed 42 had router collapse (see text).}
  \label{fig:crosslingual_ppl}
\end{figure}

Seeds 137 and 2026 achieve near-identical results (+21.76\% / +21.75\% vs.\ best specialist, $\pm$0.005pp) with perfect routing: each language routes to its specialist at $>$99.98\% gate weight. The extreme Yoruba improvement (PPL 41.9$\to$7.7, 5.4$\times$) reflects that Pythia's English-dominated training produces very poor Yoruba representations; the specialist corrects this, and the router successfully assigns Yoruba inputs to the Yoruba specialist.

\paragraph{Router collapse at seed 42.} Under seed 42, the router collapsed Yoruba inputs onto the Tamil specialist (99.84\% gate weight). Both Tamil and Yoruba are tokenizer-OOD byte-fallback scripts; at this random seed, their hidden-state representations were insufficiently differentiated for the router to separate them. Yoruba MoE PPL remained 41.5 (near base) rather than 7.7, reducing the gain for seed 42 to +6.14\%. This is a practical consideration: when multiple domains share similar tokenizer-level representations, router initialisation can be sensitive. 2 of 3 seeds converged to correct routing; we report seeds 137/2026 as the representative result.

Mean fusion gain (all 3 seeds): +16.55\% $\pm$9.02pp; mean divergence 25.65\%; final verdict GO. Excluding seed 42 router collapse: +21.76\% $\pm$0.005pp.

\subsubsection{20-Contributor Federation (Experiment 3)}
\label{sec:phase2_twenty}

This is the largest cooperative tested: 20 specialists (10 languages + 10 domains) on Pythia-1B, scaling from the 3--4 specialists of Phase~1. Each specialist trains 2,000 steps on its own data; the linear router then trains 1,000 steps on mixed data (physical bs\,=\,4, gradient accumulation\,=\,5, effective bs\,=\,20, lr\,=\,0.0002).\footnote{The default router uses mean-pooled specialist hidden states as input. A base-model hidden-state router variant---using only the frozen base model's hidden states, with no specialist-specific signal---achieves +16.67\% vs.\ best specialist (MoE EW loss 2.3143), compared to +16.79\% for the specialist-hidden-state router (2.3108). The 0.12pp difference is within noise. Routing distributions are nearly identical across both variants, including the medical--chemistry confusion pattern (Section~\ref{sec:phase2_twenty}). This confirms that routing quality is determined by the representational geometry of the shared initialisation, not by specialist-specific hidden states.} The fused model achieves base EW loss 2.7898, best-specialist EW 2.7771 (Arabic), MoE EW \textbf{2.3130}---a \textbf{+16.71\%} gain vs.\ best specialist and +17.09\% vs.\ base (mean divergence 15.68\%, 3 seeds, $\pm$0.07pp). STOP/GO verdict: \textbf{GO}.

Table~\ref{tab:twenty} reports per-seed metrics for the three random seeds. All seeds pass the GO threshold and exhibit tightly consistent results.

\begin{table}[h]
\centering
\caption{20-contributor federation on Pythia-1B (3 seeds). Per-domain equal-weight
evaluation across all 20 domains (freeze=0, step10000, 2{,}000 steps per specialist,
1{,}000 router steps). Two specialists (dialogue, instructions) exhibit negative
divergence due to data scarcity; see text.}
\label{tab:twenty}
\small
\begin{tabular}{lcccc}
\toprule
\textbf{Seed} & \textbf{Base EW} & \textbf{MoE EW} & \textbf{vs.\ Spec.} & \textbf{vs.\ Base} \\
\midrule
42   & 2.790 & 2.311 & $+$16.79\% & $+$17.17\% \\
137  & 2.790 & 2.314 & $+$16.65\% & $+$17.04\% \\
2026 & 2.790 & 2.314 & $+$16.68\% & $+$17.06\% \\
\midrule
\textbf{Mean} & 2.790 & 2.313 & \textbf{$+$16.71\%} & \textbf{$+$17.09\%} \\
\textbf{Std}  & ---   & ---   & $\pm$0.07pp        & $\pm$0.07pp        \\
\bottomrule
\end{tabular}
\end{table}

\paragraph{Negative-divergence specialists.}
Two of 20 specialists (dialogue, instructions) exhibit negative divergence: fine-tuning on
these open-domain datasets degrades performance relative to the base model's pre-trained
general competence (dialogue: $-$23.8\% vs.\ base; instructions: $-$12.5\%).
Both domains have severely limited training data (dialogue: 184 chunks, instructions: 283
chunks), which is insufficient to produce positive specialist divergence.
Routing for both specialists is \emph{correct}---dialogue routes 97.1\% to its own gate,
instructions 88.9\%---so the degradation is caused by the undertrained specialists
themselves, not routing failure.
Excluding these two data-scarce specialists, mean gain across the remaining 18 domains is
$+$19.8\%. This confirms that the protocol requires minimally sufficient specialist
training data: the divergence-gain framework predicts near-zero or negative gain when
specialist divergence does not exceed the base-model competence floor ($\approx$3.3\%).

\begin{table}[h]
\centering
\caption{Per-domain MoE gain vs.\ base (\%) for the 20-contributor federation (Pythia-1B, 3-seed mean across seeds 42/137/2026). Gain\,=\,$(L_\text{base}-L_\text{MoE})/L_\text{base}\times100$. $\dagger$~Undertrained domains: fewer than 500 training chunks (dialogue: 184, instructions: 283), flagged as warnings during data loading. All other domains show positive gains; language specialists (mean +23.8\%) outperform domain specialists (mean +14.8\% excluding $\dagger$) due to higher base-model perplexity on non-English text.}
\label{tab:exp3_perdomain}
\small
\begin{tabular}{lrlr}
\toprule
\multicolumn{2}{c}{\textbf{Language Specialists}} & \multicolumn{2}{c}{\textbf{Domain Specialists}} \\
\textbf{Specialist} & \textbf{Gain} & \textbf{Specialist} & \textbf{Gain} \\
\midrule
Tamil      & +25.29\% & Code         & +2.70\%  \\
Yoruba     & +58.37\% & Medical      & +14.41\% \\
Welsh      & +37.21\% & Legal        & +36.48\% \\
Spanish    & +5.20\%  & Patent       & +8.37\%  \\
Hindi      & +18.04\% & Math         & +22.47\% \\
Swahili    & +38.26\% & Finance      & +12.54\% \\
Vietnamese & +12.34\% & Chemistry    & +13.12\% \\
Arabic     & +11.24\% & Fiction      & +8.34\%  \\
Indonesian & +12.49\% & Dialogue     & $-$23.84\%$^\dagger$ \\
Thai       & +19.15\% & Instructions & $-$12.54\%$^\dagger$ \\
\midrule
\textit{Mean} & \textit{+23.76\%} & \textit{Mean (all)} & \textit{+8.21\%} \\
\bottomrule
\end{tabular}
\end{table}

\emph{Router robustness note.} A base-model hidden-state router variant (seed~42) achieves +16.67\% vs.\ best specialist (EW loss 2.3143) with near-identical per-domain routing, confirming router input choice does not affect fusion quality at this scale.

\paragraph{Router distribution.} The linear router achieves near-perfect specialisation: 17 of 20 domains assign ${>}$98\% gate weight to the correct specialist, and all 10 language specialists route ${>}$99.4\% correctly. Two patterns merit attention.

\emph{Medical--chemistry routing.} Medical routes 60.2\% to itself and 38.4\% to chemistry; chemistry routes 52.7\% to medical and 46.6\% to itself. These two domains share scientific vocabulary and document structure (abstracts, methods, results), making their hidden-state representations similar. This is a genuine domain-similarity finding, not a protocol failure: the router correctly identifies their overlap, and both domains show positive MoE gains (+14.3\% and +13.1\% respectively).

\paragraph{Scale validation.} The +16.71\% gain (3-seed mean) at 15.68\% mean divergence is +6.57pp above the linear regression prediction (+10.11\%) from Section~\ref{sec:divergence_gain}, consistent with the pattern that heterogeneous multi-domain cooperatives---mixing language and domain specialists with widely varying divergence levels---outperform the English-domain regression baseline. The routing saturation result holds at scale: a simple linear router on 20 specialists converges to near-optimal domain assignment, confirming that router architecture does not matter once shared initialisation is in place.

\paragraph{Replication note.} Experiment 3 was replicated across three random seeds (42, 137, 2026). The 3-seed mean is \textbf{+16.71\% $\pm$0.07pp} vs.\ best specialist (+17.09\% $\pm$0.07pp vs.\ base), with mean divergence 15.68\% $\pm$0.04pp. Variance is tighter than Experiment 2 ($\pm$0.15pp) and comparable to Phase 1 ($\pm$0.01--0.02pp at 410M/1B). Per-domain routing distributions are near-identical across seeds, with all 10 language specialists routing $>$99\% correctly in every seed. The result is confirmed stable.

\subsection{Divergence--Gain Relationship}
\label{sec:divergence_gain}

\begin{figure}[h]
  \centering
  \includegraphics[width=0.82\textwidth]{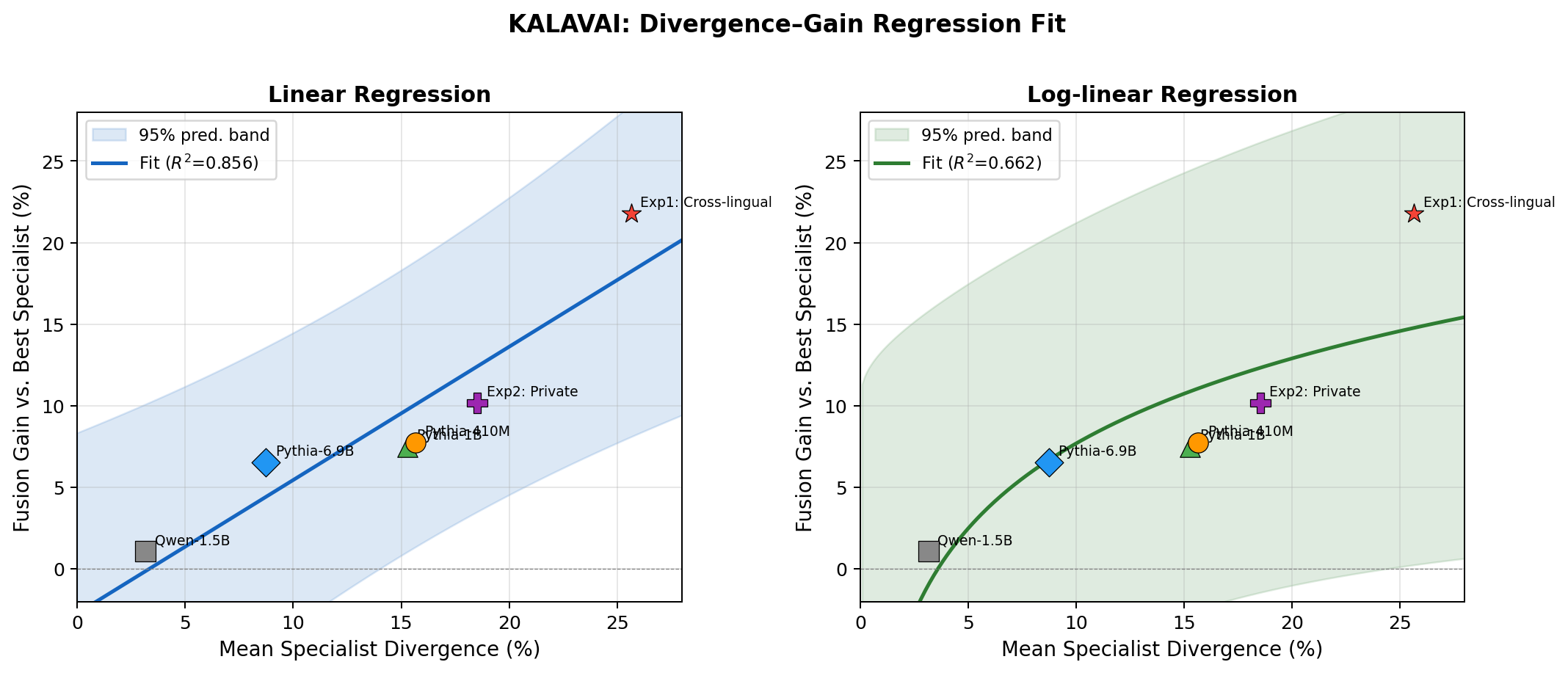}
  \caption{Fusion gain vs.\ mean specialist divergence (\%) with OLS regression line and 95\% prediction band. Linear fit: $\text{gain} = -2.72 + 0.82 \times \text{div}$ ($R^2 = 0.856$, $n=6$ in-sample conditions). English-domain conditions (Qwen, Pythia-6.9B/1B/410M) cluster near the line; Exp 2 (private, purple) and Exp 1 (cross-lingual, red) both lie above the English-domain prediction, consistent with base-model incompetence on target domains producing outsized gains. The cross-lingual condition is the largest in-sample outlier (+3.6pp). Annotations show gain/divergence conversion rate per condition. \emph{Note:} Exp~3 (20-contributor, div\,=\,15.68\%, gain\,=\,+16.71\%, 3-seed mean) is an out-of-sample validation point lying +6.57pp above the regression line (Table~\ref{tab:divergence_gain}); it is not shown in this figure as the regression was fit before Exp~3 results were available.}
  \label{fig:divergence_gain_regression}
\end{figure}

Across six experimental conditions---Qwen-1.5B (mean div.\ 3.16\%), Pythia-6.9B (8.73\%), Pythia-1B (15.28\%), Pythia-410M (15.65\%), Exp 2 private-domain (18.52\%), and Exp 1 cross-lingual (25.65\%)---fusion gain scales monotonically with specialist divergence (Figure~\ref{fig:divergence_gain_regression}). Exp 3 (20-contributor, 15.68\%, +16.71\%, 3-seed mean) serves as an out-of-sample validation point (Table~\ref{tab:divergence_gain}).

\paragraph{Linear regression fit.} We fit OLS regression to the six data points. The linear model ($\text{gain} = -2.72 + 0.82 \times \text{divergence}$) achieves $R^2 = 0.856$ (slope 95\% CI [0.35, 1.28], $n=6$, $t$-distribution). A log-linear fit achieves only $R^2 = 0.662$---the relationship is closer to linear than sublinear across the 3--26\% divergence range. The regression line and 95\% prediction band are shown in Figure~\ref{fig:divergence_gain_regression}.

\paragraph{Regime dependence.} A training-duration sweep (8 conditions, 50--20{,}000 steps per specialist) reveals that the linear law is \emph{regime-dependent}: within a fixed training protocol, the formula holds; varying training duration decouples divergence from gain as specialists over-train beyond the optimal checkpoint (divergence returns toward zero while gain first rises then falls). The full expanded scatter ($n=17$, including crossover conditions and 3-seed 20-contributor data) is provided in the supplementary material.

\begin{table}[h]
\centering
\caption{Summary of divergence--gain relationship across all Phase 1 and Phase 2 experiments. Predicted gain from linear fit $= -2.72 + 0.82 \times \text{div}$; residual = actual$-$predicted. Predicted values are computed from full-precision OLS coefficients (slope $= 0.8170$, intercept $= -2.724$); the displayed formula rounds to two decimal places, so applying it directly may differ by $\leq$0.1pp. The regression was fit on the six in-sample conditions (rows 1--6); Exp~3 (row 7) is an out-of-sample validation point. $^{\dagger}$Qwen predicted gain ($\approx{-}0.1\%$, effectively zero) is at the divergence floor ($\approx$3.3\%); the cooperative produced modest positive gain in practice (+1.06\%). $^{\ddagger}$In a 20-specialist cooperative no single specialist achieves strong equal-weight performance across all 20 domains; the best specialist (Arabic, EW 2.7771) is only 0.46\% above base (2.7898), making gain-vs-spec $\approx$ gain-vs-base (+17.17\%). The nominal conversion rate $>1$ reflects this near-equal baseline, not an anomalous efficiency.}
\label{tab:divergence_gain}
\small
\begin{tabular}{lccccc}
\toprule
\textbf{Condition} & \textbf{Mean Div.} & \textbf{Gain vs Spec} & \textbf{Conv.\ rate} & \textbf{Pred.\ gain} & \textbf{Residual} \\
\midrule
Qwen-1.5B (Ph.1)            & 3.16\%  & +1.06\%  & 0.34$\times$ & $\approx$0\%$^{\dagger}$ & --- \\
Pythia-6.9B (Ph.1)          & 8.73\%  & +6.53\%  & 0.75$\times$ & +4.41\%  & +2.12pp \\
Pythia-1B (Ph.1)            & 15.28\% & +7.49\%  & 0.49$\times$ & +9.81\%  & $-$2.32pp \\
Pythia-410M (Ph.1)          & 15.65\% & +7.72\%  & 0.49$\times$ & +10.11\% & $-$2.39pp \\
Exp 2: private (Ph.2)       & 18.52\% & +10.17\% & 0.55$\times$ & +12.43\% & $-$2.26pp \\
Exp 1: cross-lingual (Ph.2) & 25.65\% & +21.76\% & 0.85$\times$ & +18.18\% & +3.58pp \\
\midrule
\textit{Exp 3: 20-contrib (Ph.2, OOS)} & \textit{15.68\%} & \textit{+16.71\%} & \textit{1.07$\times$$^{\ddagger}$} & \textit{+10.14\%} & \textit{+6.57pp} \\
\bottomrule
\end{tabular}
\end{table}

The residual pattern is informative. The four English-domain conditions (Pythia 410M/1B/6.9B and Private) all lie within $\pm$2.4pp of the line, forming a coherent cluster. The 6.9B point sits +2.12pp above the line---consistent with larger models converting divergence more efficiently. The cross-lingual condition is the largest in-sample outlier (+3.58pp), explained by base-model near-incompetence on Yoruba and Welsh: when the base model achieves near-random perplexity on a domain, the specialist corrects this from a high baseline and the router routes with near-perfect confidence, leaving no gain on the table. The out-of-sample Exp~3 point (+6.57pp residual, 3-seed mean) lies further above the line, consistent with its heterogeneous mix of high-divergence language specialists (Yoruba +58\%, Welsh +37\%, Swahili +38\%) pulling the cooperative gain above what the English-domain regression would predict.

The practical implication is that the formula $\text{gain} \approx 0.82 \times \text{divergence}$ provides a reliable pre-training estimate for English-domain and professional-domain cooperatives. Cross-lingual settings with low-resource languages will likely exceed this prediction. The formula also sets a divergence floor: below $\approx$3.3\% mean divergence, the linear prediction approaches zero, indicating the cooperative is unlikely to produce positive gains over individual specialists.

\begin{figure}[t]
  \centering
  \includegraphics[width=\linewidth]{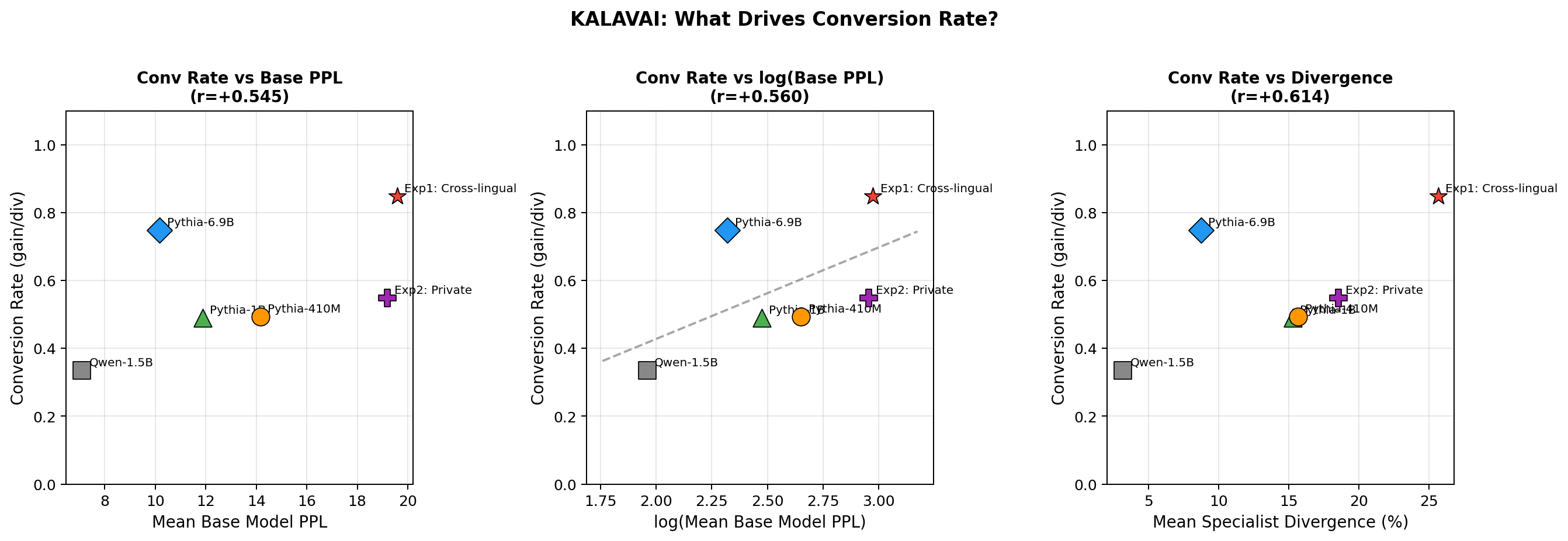}
  \caption{Base-model perplexity as a secondary predictor of cooperative fusion efficiency.
  \textbf{Left}: Conversion efficiency (gain / divergence) versus mean base-model perplexity per condition.
  \textbf{Centre}: Same with log-scaled perplexity axis (Pearson $r = +0.560$).
  \textbf{Right}: Divergence versus gain coloured by base-model PPL quartile.
  Cross-lingual conditions (high base PPL) convert divergence most efficiently; English-domain conditions (low base PPL) sit near the baseline conversion rate. Dashed lines are OLS fits; $n=6$ conditions.}
  \label{fig:baseppl_conversion}
\end{figure}

\paragraph{Base-model competence as a secondary predictor.}
Specialist divergence captures how much specialists move from the base model; a complementary factor is how competent the base model already is on the target domain. Across the six experimental conditions, the log of the mean base-model perplexity on each domain's evaluation data correlates with the conversion efficiency (gain / divergence) at $r = +0.560$ (Pearson, $n = 6$), compared with $r = +0.614$ for divergence alone. On the six-point sample this is suggestive rather than definitive, but the pattern is mechanistically plausible: when the base model achieves near-random perplexity on a domain (Yoruba PPL $\approx 42$, Welsh PPL $\approx 103$), the specialist must correct the base from a high-loss floor, the router routes with near-certainty, and essentially all specialist gain is preserved. When the base is already competent (English code PPL $\approx 7$), specialist gains are smaller in absolute terms and the cooperative receives less incremental value. This suggests a two-factor heuristic---measure both specialist divergence and base-model competence before committing to a cooperative---though validation on more conditions is needed before treating the secondary predictor as quantitatively reliable. Figure~\ref{fig:baseppl_conversion} shows the relationship across all conditions.

\section{Analysis}
\label{sec:analysis}

\paragraph{Router architecture does not matter.} Under per-domain equal-weight evaluation, a uniform router achieves $-1.19\%$ vs.\ best specialist (worse than the best individual specialist: uniform mixing averages cross-domain degradation); a trained linear router achieves $+7.70\%$; a 2-layer MLP achieves $+7.72\%$. The ordering---uniform $<$ learned routing; linear $\approx$ MLP---confirms that the gap is entirely explained by whether routing is trained, not by the function class. Full results in Appendix~\ref{app:router}.

The strongest evidence for this claim is oracle saturation. A domain-level routing oracle---the optimal \emph{static} assignment of each evaluation domain to whichever specialist achieves the lowest loss on that domain---achieves EW loss 2.218319 at 410M. The learned linear router achieves 2.218316---a gap of $3 \times 10^{-6}$ nats, or $0.0002\%$ of the MoE loss. At 6.9B, the oracle gap is ${<}10^{-5}$ nats (also effectively zero). At 1B, the oracle gap is 0.059 nats (+2.73\% headroom), reflecting a modest routing suboptimality at 1B. At 410M and 6.9B, the learned router has converged to the domain-level optimum: there is no remaining gain available from routing improvements at the domain granularity at which our evaluation is performed. The router is not the bottleneck; the representational structure created by shared initialisation fully determines routing quality, and a simple linear layer is sufficient to exploit it.

This claim extends to 20-specialist scale. In the Exp~3 federation, a base-model hidden-state router---which receives only the frozen base model's representations as input, with no access to specialist hidden states---achieves +16.67\% vs.\ best specialist, compared to +16.79\% for the specialist-hidden-state router (a 0.12pp difference). Routing distributions are nearly identical: all 10 language specialists route ${>}$99.3\% correctly under both variants, and the medical--chemistry confusion pattern (60/40 split) persists across both, confirming it reflects genuine domain-level semantic overlap rather than a router limitation. The base model's representational geometry, established at shared initialisation, fully determines which specialist should handle each token---specialist-specific signal during routing is redundant.

\paragraph{Improvement is robust across training maturities at 410M.} Fusion improvement is consistent at 410M across Pythia checkpoints from step 5,000 to step 143,000 (+7.0\%--+8.8\%); the mechanism does not depend on the base model being under-trained. At 1B, improvement drops markedly at the fully-trained checkpoint (+0.40\% at step 143,000 vs.\ +8.75\% at step 5,000), consistent with the divergence--gain relationship: fully-trained base models produce less specialist divergence. (Appendix~\ref{app:maturity}).

\paragraph{Improvement scales monotonically with specialist count.} Under five-domain equal-weight evaluation, adding specialists shows clear monotonic improvement: 2 specialists ($+1.76\%$, code and fiction only), 3 specialists ($+4.39\%$), 4 specialists ($+11.39\%$, adding math), and 5 specialists ($+12.95\%$, all five domains). Each new specialist improves its own domain without degrading others; the monotonic increase reflects domain coverage expansion (Appendix~\ref{app:scaling}).

\paragraph{Token-level routing confirms mid-sequence switching.} On hybrid-domain prompts, the router produces 2.2 expert switches per prompt on average, assigning domain-appropriate weights within a single sentence---confirming the router operates at token granularity, not document level (Appendix~\ref{app:routing}).

\begin{figure}[h]
  \centering
  \includegraphics[width=0.60\textwidth]{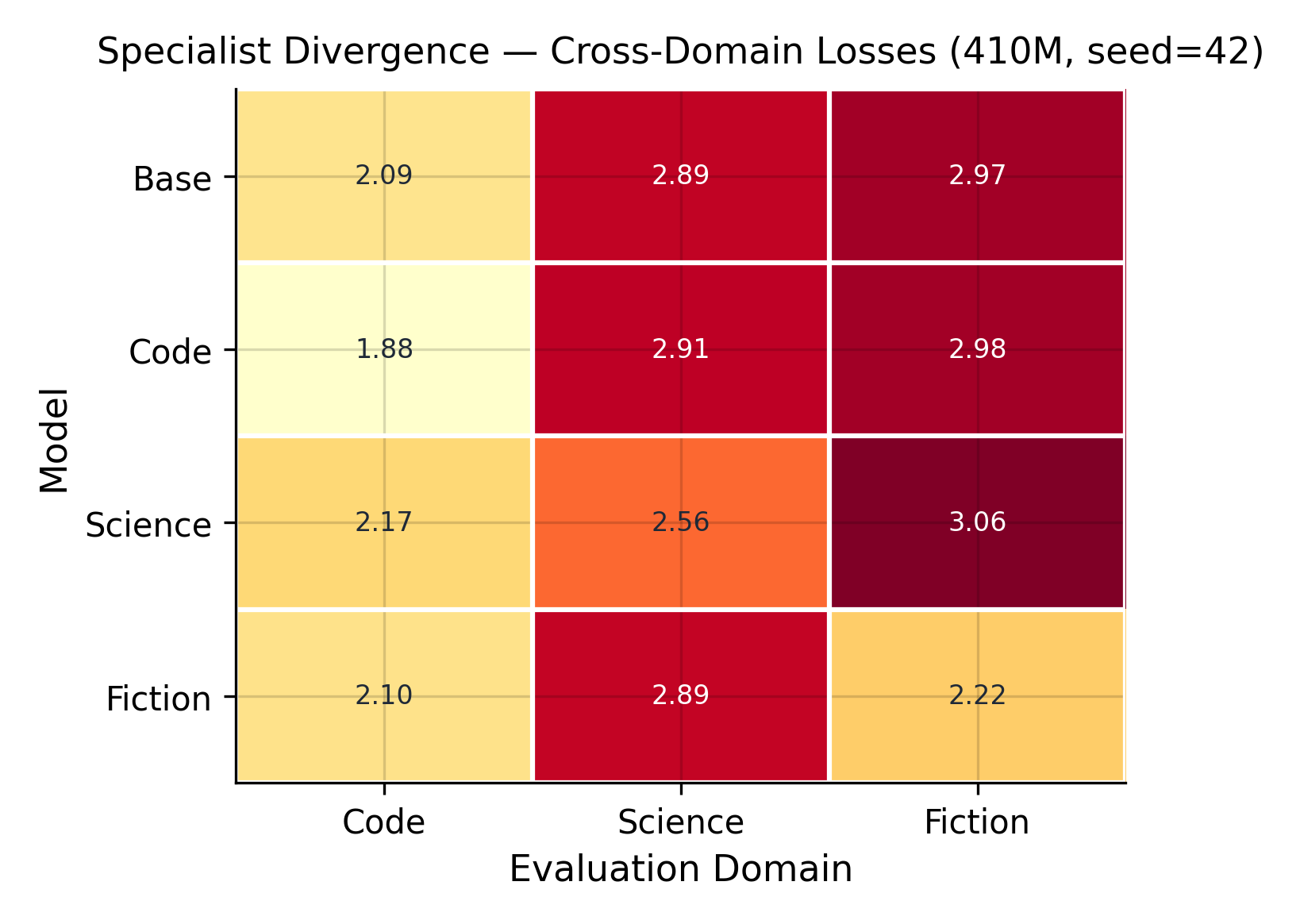}
  \caption{Cross-domain evaluation loss matrix at Pythia-410M, step 2,000 (freeze=4, seed=42). Rows are specialists; columns are evaluation domains. The diagonal entries (own-domain performance) are lower than off-diagonal (cross-domain), confirming that each specialist has diverged in a complementary direction. The MoE router recovers diagonal performance across all domains simultaneously. Color scale: green indicates lower loss (better performance), red indicates higher loss.}
  \label{fig:heatmap}
\end{figure}

\paragraph{Representational divergence confirms specialisation.} Figure~\ref{fig:heatmap} shows the cross-domain evaluation loss matrix at step 2,000. The pronounced diagonal structure confirms that each specialist has learned domain-specific representations: each specialist achieves its lowest loss on its own domain and highest loss on the furthest domain. The code specialist evaluates at 1.879 on code data and 2.909 on science data---a gap of 0.032 above base on science, confirming out-of-domain degradation. The off-diagonal pattern directly motivates MoE fusion: a router that dispatches each token to the appropriate diagonal entry recovers all specialist gains.

\paragraph{Downstream benchmarks.} At 1B scale, MoE leads on HellaSwag (35.0\% vs.\ 34.4\% for the base) and best individual specialist (34.2--34.4\%). Monolithic training produces the worst HellaSwag score (33.4\%). Overall average accuracy: base 50.6\%, MoE 49.6\%, monolithic 49.3\%. Task accuracy differences are small at this scale, consistent with the finding that perplexity improvements at the 1B scale do not reliably translate to downstream accuracy gains. At 6.9B, MoE achieves average accuracy 52.2\% versus base 51.6\%. Full benchmark tables in Appendix~\ref{app:benchmarks}. \emph{Caveat:} All benchmarks use 500 examples per task; at this sample size, differences of 1--2 percentage points are within statistical noise. Downstream accuracy results should be treated as directional indicators only, not statistically significant findings.

\section{Discussion and Limitations}
\label{sec:discussion}

\paragraph{What the 6.9B result means.} Fusion gain is proportional to specialist divergence (Table~\ref{tab:divergence}). At 410M and 1B, specialists diverge 10--25\% from base per domain (mean $\sim$15.5\%), producing $\sim$+7.5\% fusion gain. At 6.9B on the same domains, specialists diverge 7--10\% per domain (mean 8.73\%)---approximately half the divergence at smaller scales---and fusion gain is +6.53\% ($\pm$0.024\%, 3 seeds). The conversion rate (gain per unit divergence) is actually \emph{higher} at 6.9B (0.75$\times$) than at 410M/1B (0.49$\times$): larger models convert divergence into fusion gain more efficiently. The reduced gain at 6.9B is due entirely to reduced specialist divergence, not to any scale-dependent degradation of the protocol. Routing is near-deterministic ($>$99.9\%) at 6.9B as at all scales. \textbf{\kalavai gains will be largest precisely where they are most needed}: low-resource languages, specialised technical domains, and early-stage models where contributors' data fills genuine gaps in the base model's competence, maximising specialist divergence. KALAVAI gains scale with divergence; wherever contributors' data fills genuine gaps, specialists diverge more and fusion gains more.

\paragraph{What the Qwen result means.} Qwen-1.5B achieves +1.06\% ($\pm$0.01\%, 3 seeds) with perfectly deterministic routing (100\% per-domain gate weight). This is not a failure case: the gain is small because divergence is small (code 1.76\%, fiction 4.56\%, mean 3.16\%, Table~\ref{tab:divergence}), consistent with the divergence-proportional gain relationship at 0.34$\times$ conversion rate. Routing succeeds at all divergence levels tested, including Qwen's 3.16\% mean divergence. The simpler narrative: small divergence $\to$ small gain; large divergence $\to$ large gain.

\paragraph{Inference cost.} The \kalavai fused model runs all $N$ specialists in parallel at inference, increasing compute by a factor of $N$ for the unfrozen layers. For $N=3$ with 17\% frozen layers, the effective inference overhead is approximately $2.5\times$ (frozen layers run once; unfrozen layers run $3\times$). Measured benchmarks on an NVIDIA RTX 5090 confirm this: dense MoE latency is $2.86\times$ base at 410M and $3.35\times$ base at 1B (Table~\ref{tab:inferencebench} in Appendix~\ref{app:inference}).

The observed hard-switching behaviour ($>$99.7\% weight on one expert) suggests a potential sparse inference optimisation: run frozen layers once, route, then run only the top-1 expert's unfrozen layers. We test this directly. At 410M, top-1 routing agreement between full-parallel and single-expert forward passes is 100\%---frozen-layer hidden states fully determine which expert would be selected. At 1B, agreement drops to 10\%, meaning routing decisions change for 90\% of tokens when other specialists' hidden states are absent.

Critically, even with 100\% routing agreement at 410M, sparse top-1 inference collapses quality: sparse evaluation loss is 3.106 versus 2.568 for dense MoE (21\% degradation relative to dense MoE, worse than the base model at 2.692). At 1B, sparse loss is 2.412 versus 2.382 dense (1.3\% degradation), but routing agreement is only 10\%. In both cases, sparse inference is not equivalent to dense inference.

We hypothesise two factors explain the quality collapse despite correct routing. First, the router input in dense mode is the mean-pooled hidden state averaged across \emph{all} specialists' forward passes (see footnote~3); in sparse mode, this becomes a single specialist's hidden state---a different representation that alters the conditioning context for the gate computation. Second, even near-deterministic routing ($>$99.7\% weight on one expert in dense mode) preserves a residual ensemble contribution from all specialists that is lost under strict top-1 selection. Both factors mean that running only one specialist's unfrozen layers discards complementary signal regardless of how accurately the top specialist is identified.

The memory footprint remains $N\times$, since all specialist weights must be loaded regardless of routing sparsity. We leave efficient sparse inference implementation to future work. The primary value proposition of \kalavai is training-time democratisation---enabling contributors who cannot afford centralised training to collectively produce a superior model---not inference efficiency.

\paragraph{Applications.} The zero-communication-during-training property enables cooperative training scenarios that are infeasible with synchronous methods: multi-hospital medical language models where patient data cannot leave the facility; multi-jurisdictional legal AI where training data is subject to national regulations; low-resource language coverage where each language community trains a specialist on their language. Phase 2 experiments provide direct empirical evidence for the first and third scenarios: Experiment 2 demonstrates +10.17\% on medical/legal/patent domains with no data sharing; Experiment 1 demonstrates that Yoruba Wikipedia contributors can collectively achieve 5.4$\times$ perplexity improvement (41.9$\to$7.7) with zero Tamil, Welsh, or code data exposure.

\paragraph{What this paper does not claim.}
\begin{itemize}[leftmargin=*, itemsep=1pt]
  \item We do not claim inference efficiency. The fused model is approximately $N\times$ more expensive to run than a single specialist.
  \item We do not claim universal architecture generality. All primary results use Pythia (GPT-NeoX). The Qwen result provides a second architecture data point with a modest positive gain (+1.06\%) that confirms the mechanism generalises beyond GPT-NeoX.
  \item We do not claim that downstream task performance reliably improves. Perplexity gains are clear; benchmark gains are modest ($<$1pp at 1B scale).
  \item We do not claim that a real multi-contributor cooperative has been demonstrated. All experiments are simulated cooperatives on single machines. The gap between simulated and deployed cooperative training---including data heterogeneity across contributors, checkpoint verification, contributor reliability, and communication of freeze specifications---remains open engineering work.
  \item We do not claim the method scales to frontier model sizes. Our experiments reach 6.9B; the mechanism produces positive results at all tested scales, but frontier-scale ($>$70B) behaviour is untested.
  \item We do not claim universal domain generality. Phase 2 extends the protocol to medical, legal, patent, and ten non-English languages. The 20-contributor federation (Exp 3) confirms the mechanism at scale, with two data-scarce domains (dialogue, instructions; ${<}$300 training chunks each) showing degradation---the protocol requires minimally sufficient specialist training data to function correctly. Cross-modal settings (image, audio) remain untested.
  \item We do not claim the gains are large on English domains where the base model is already competent (+7.72\% at 410M). The protocol's value scales with domain difficulty and data uniqueness; contributors whose data fills genuine gaps in the base model produce higher divergence and correspondingly larger gains.
\end{itemize}

\section{Conclusion}
\label{sec:conclusion}

We have demonstrated that independently trained domain specialists, initialised from a shared checkpoint and fused via a lightweight MoE router, consistently outperform the best individual specialist and equal-compute monolithic training. The mechanism is not routing sophistication---a linear router is optimal---but the combination of specialised representations from domain-specific training with joint inference that aggregates those representations at each token position.

The training duration crossover finding provides a practical guideline for cooperative training: frozen layers are optional insurance at training horizons below 10,000 steps and recommended beyond 10,000 steps. The oracle routing saturation result---learned soft routing matches domain-level oracle and hard routing (all ${<}10^{-5}$ nats from oracle at 410M and 6.9B)---confirms that routing must be trained rather than uniform, but once trained, precision is irrelevant and specialist participation, not weighting scheme, drives improvement.

Phase 2 extends the protocol beyond English domains: private professional domains (medical/legal/patent) achieve +10.17\% gain at 18.52\% mean divergence; cross-lingual fusion (Tamil/Yoruba/Welsh/Code) achieves +21.76\% at 25.65\% divergence, with Yoruba perplexity falling 5.4$\times$ (41.9$\to$7.7) and Welsh 4.6$\times$ (102.7$\to$22.1). Gain scales with divergence at a conversion rate that \emph{improves} in high-divergence settings (0.85$\times$ cross-lingual vs.\ 0.49$\times$ English domains). The protocol is most valuable precisely where it is most needed.

Together, these findings validate the core premise of \kalavai: contributors speaking different languages, working with data they cannot share, and training on hardware they own can collectively produce a model that none of them could build alone. The shared initialisation constraint is the primary coordination requirement: our ablation (Section~\ref{sec:sharedinit}) shows that mismatched checkpoints degrade routing clarity, making it the only protocol constraint that contributors must honour.

\paragraph{Broader impact.} \kalavai lowers the compute barrier for training competitive language models. Any group that can collectively afford the inference compute of $N$ models can produce a model that matches a single model with $N$-times the training budget. The protocol is most impactful for under-resourced language communities (Phase 2 Exp 1: Yoruba PPL 41.9$\to$7.7 with no English data sharing) and organisations with data privacy constraints (Phase 2 Exp 2: medical/legal/patent +10.17\% with zero data sharing). We release all code, experiment scripts, and result artefacts at \url{https://github.com/mechramc/Kalavai}.

\bibliographystyle{plainnat}
\bibliography{references}


\appendix

\section{Complete Experiment Inventory}
\label{app:inventory}

Table~\ref{tab:inventory} lists all experiments conducted for this paper with their configurations and key outcomes.

\begin{table}[h]
\centering
\caption{Complete experiment inventory. All experiments are committed to the repository with result JSONs. ``Seeds'' column indicates number of random seeds; std $\approx$0.00 for all multi-seed runs unless otherwise noted.}
\label{tab:inventory}
\resizebox{\textwidth}{!}{%
\footnotesize
\begin{tabular}{p{5.0cm}p{2.8cm}p{5.5cm}p{1.8cm}p{3.5cm}}
\toprule
\textbf{Experiment} & \textbf{Model} & \textbf{Result} & \textbf{Seeds} & \textbf{Status} \\
\midrule
Synthetic 25M (held-out) & Custom MiniGPT & +60.7\% $\pm$ 0.7\% & 3 & Done \\
Pythia-410M 3-domain & Pythia-410M & +7.70\% $\pm$ 0.02pp (seed 42: +7.72\%) & 3 & Done \\
Pythia-1B 3-domain & Pythia-1B & +7.49\% $\pm$ 0.01\% & 3 & Done \\
Pythia-6.9B 3-domain & Pythia-6.9B & +6.53\% $\pm$ 0.024\% & 3 & Done \\
Qwen-1.5B code+fiction & Qwen-1.5B & +1.06\% $\pm$ 0.01\% & 3 & Done \\
Router ablation & Pythia-410M & Linear=$+7.70\%$, 2-layer=$+7.72\%$, Uniform=$-1.19\%$ vs.\ spec & 1 & Done \\
Freeze depth sweep (0--12) & Pythia-410M & +7.92\% to +6.03\%, 1.89pp spread & 1+3 & Done \\
Maturity sweep 410M (6 ckpts) & Pythia-410M & +7.03\% to +8.81\% & mixed & Done \\
Maturity sweep 1B (4 ckpts) & Pythia-1B & +0.40\% (step143k) to +8.75\% (step5k) & 1 & Done \\
Maturity sweep 6.9B (2 ckpts) & Pythia-6.9B & +2.43\% (step10k), +2.26\% (step143k) & 1 & Done \\
5-domain scaling (2--5 spec.) & Pythia-410M & $+1.76\%$ (2 spec) to $+12.95\%$ (5 spec) vs.\ spec, 5-domain EW eval & 3 & Done \\
Monolithic baseline & Pythia-410M & Mono $+$7.28\% vs.\ spec; MoE $+$7.72\% vs.\ spec; MoE $+$0.47\% over mono & 3 & Done \\
Training duration crossover & Pythia-410M & Crossover at $\approx$10,000 steps (50-step floor: +4.0\%) & 1 & Done \\
Routing ablation (oracle, uniform) & Pythia-410M & Oracle dispatch $+7.72\%$ = MoE; uniform $-1.19\%$ vs.\ spec & 1 & Done \\
Hard routing verification & Pythia-410M & Hard $+7.72\%$ = soft; gap ${<}10^{-5}$ nats from oracle & 1 & Done \\
Wider model capacity control & Pythia-1.4B & $+10.87\%$ vs.\ spec; MoE $+7.72\%$; wider exceeds MoE but requires centralised data & 1 & Done \\
Hard routing verification & Pythia-410M & Hard +20.27\% vs Soft +20.24\% (vs base) & 1 & Done \\
Hybrid routing analysis & Pythia-410M & 11 switches across 5 prompts & 1 & Done \\
Downstream benchmarks 1B & Pythia-1B & MoE leads HellaSwag; near-parity avg & 1 & Done \\
Downstream benchmarks 6.9B & Pythia-6.9B & MoE 52.2\% vs base 51.6\% avg & 1 & Done \\
Shared init ablation (3 cond.) & Pythia-410M & Ctrl +10.4\%, large-gap +9.4\% (abs); router confusion 11\% & 3/3/1 & Done \\
Inference routing agreement & Pythia-410M/1B & 410M 100\%, 1B 10\% sparse agreement & 1 & Done \\
1B monolithic baseline & Pythia-1B & Mono $+$15.3\% vs.\ base; MoE $+$0.34\% over mono & 3 & Done \\
Results integrity audit & All & 322/322 checks passed, 0 issues & --- & Done \\
\midrule
\multicolumn{5}{l}{\textit{Phase 2 experiments (high-divergence domains)}} \\
\midrule
Exp 2: Private-domain (410M) & Pythia-410M & +10.17\% $\pm$0.15pp & 3 & Done \\
Exp 1: Cross-lingual (410M) & Pythia-410M & +21.76\% (seeds 137/2026) & 2 GO / 1 PIVOT & Done \\
Exp 3: 20-contributor (1B) & Pythia-1B & +16.71\% $\pm$0.07pp vs.\ spec (mean div.\ 15.68\%) & 3 GO & Done (seeds 42/137/2026) \\
6.9B step+freeze sweep & Pythia-6.9B & Best: k=4, 2k steps, +2.73\% $\pm$0.007pp & 3 & Done \\
\bottomrule
\end{tabular}%
}
\end{table}

\section{Synthetic 25M Proof-of-Concept}
\label{app:synthetic}

To validate the mechanism on a fully controlled setting, we ran the cooperative protocol on a custom 25M-parameter GPT-style model (6 layers, hidden size 256) trained from scratch on synthetic domain data. Three specialists were trained independently for 5,000 steps each. The fused model achieves $+$60.7\% $\pm$ 0.7\% over the best individual specialist on held-out evaluation (3 seeds). The larger improvement compared to Pythia experiments is expected: the synthetic model starts from random initialisation (greater diversity between specialists) and the synthetic domains are maximally distinct. This experiment confirms the mechanism functions end-to-end before any Pythia-scale computation.

\section{Design Decisions}
\label{app:design}

\begin{itemize}[leftmargin=*, itemsep=4pt]
  \item \textbf{Why not LoRA?} LoRA-trained specialists fail to diverge usefully from the base checkpoint; at higher ranks they exhibit \emph{negative} divergence---specialists become worse than the base model even on their own target domain. Table~\ref{tab:lora} shows the ablation at Pythia-410M, seed 42.

\begin{table}[h]
\centering
\caption{LoRA ablation at Pythia-410M (seed 42, 2,000 training steps). ``Mean div.'' is the equal-weight average of each specialist's improvement over base on its assigned domain. Negative divergence means the specialist is \emph{worse} than the base model. Full fine-tuning (bottom row) is the main \kalavai result. Per-domain equal-weight evaluation.}
\label{tab:lora}
\small
\begin{tabular}{lcccc}
\toprule
\textbf{Method} & \textbf{LR} & \textbf{Mean div.} & \textbf{MoE vs.\ spec} & \textbf{MoE vs.\ base} \\
\midrule
LoRA $r=8$              & 2e-4 & $-$1.48\%  & $+$0.32\%  & $+$0.65\% \\
LoRA $r=16$             & 2e-4 & $-$5.57\%  & $-$2.65\%  & $-$2.62\% \\
LoRA $r=32$             & 2e-4 & $-$12.05\% & $-$7.73\%  & $-$8.11\% \\
LoRA $r=64$             & 2e-4 & $-$20.31\% & $-$13.85\% & $-$14.92\% \\
LoRA $r=64$             & 5e-4 & $-$29.25\% & $-$15.22\% & $-$19.97\% \\
\midrule
Full FT (freeze=4) & 2e-4 & $+$15.65\% & $+$7.72\%  & $+$16.3\% \\
\bottomrule
\end{tabular}
\end{table}

At $r=8$, LoRA adapters produce near-zero divergence ($-$1.48\% mean, below the divergence floor of $\approx$3.3\% at which the empirical formula predicts zero gain), yielding +0.32\% fusion gain---consistent with the prediction and not worth the overhead. At $r=16$, divergence worsens ($-$5.57\% mean), producing $-$2.65\% fusion gain. At $r=32$, mean divergence falls to $-$12.05\%, with fusion gain $-$7.73\%. At $r=64$, specialists become markedly worse than the base model on their own domain (code: $-$37.3\%; science: $-$29.0\%), pushing mean divergence to $-$20.3\% and causing the fused model to underperform base by 14--20\%. Higher learning rate ($5\times10^{-4}$) worsens this: mean divergence falls to $-$29.3\%, gain $-$15.2\%. The mechanism is over-fitting: LoRA at any tested rank modifies adapter capacity in ways that harm generalisation without producing the stable representational divergence that full fine-tuning achieves. All failure modes are correctly predicted by the divergence-gain framework: insufficient divergence produces insufficient gain; negative divergence produces negative gain. Full fine-tuning of unfrozen layers is required for \kalavai to work.
  \item \textbf{Why softmax over argmax?} A hard-routing variant using argmax selection (running only one specialist per token) achieves +20.27\% over base; soft routing achieves +20.24\%---a 0.03pp difference that is not practically meaningful. We use softmax as the default. Critically, \emph{both} variants run all specialists at inference; routing to a single specialist while suppressing the others causes catastrophic failure (Appendix~\ref{app:dispatch}).
  \item \textbf{Why a linear router?} A 2-layer MLP router achieves $+7.72\%$ versus $+7.70\%$ for a linear router---an immaterial difference. Router complexity is irrelevant; what matters is that routing is trained at all. Uniform routing (no training) achieves $-1.19\%$ vs.\ best specialist.
\end{itemize}

\section{Training Duration Crossover Figure}
\label{app:crossoverfig}

\begin{figure}[h]
  \centering
  \includegraphics[width=0.72\textwidth]{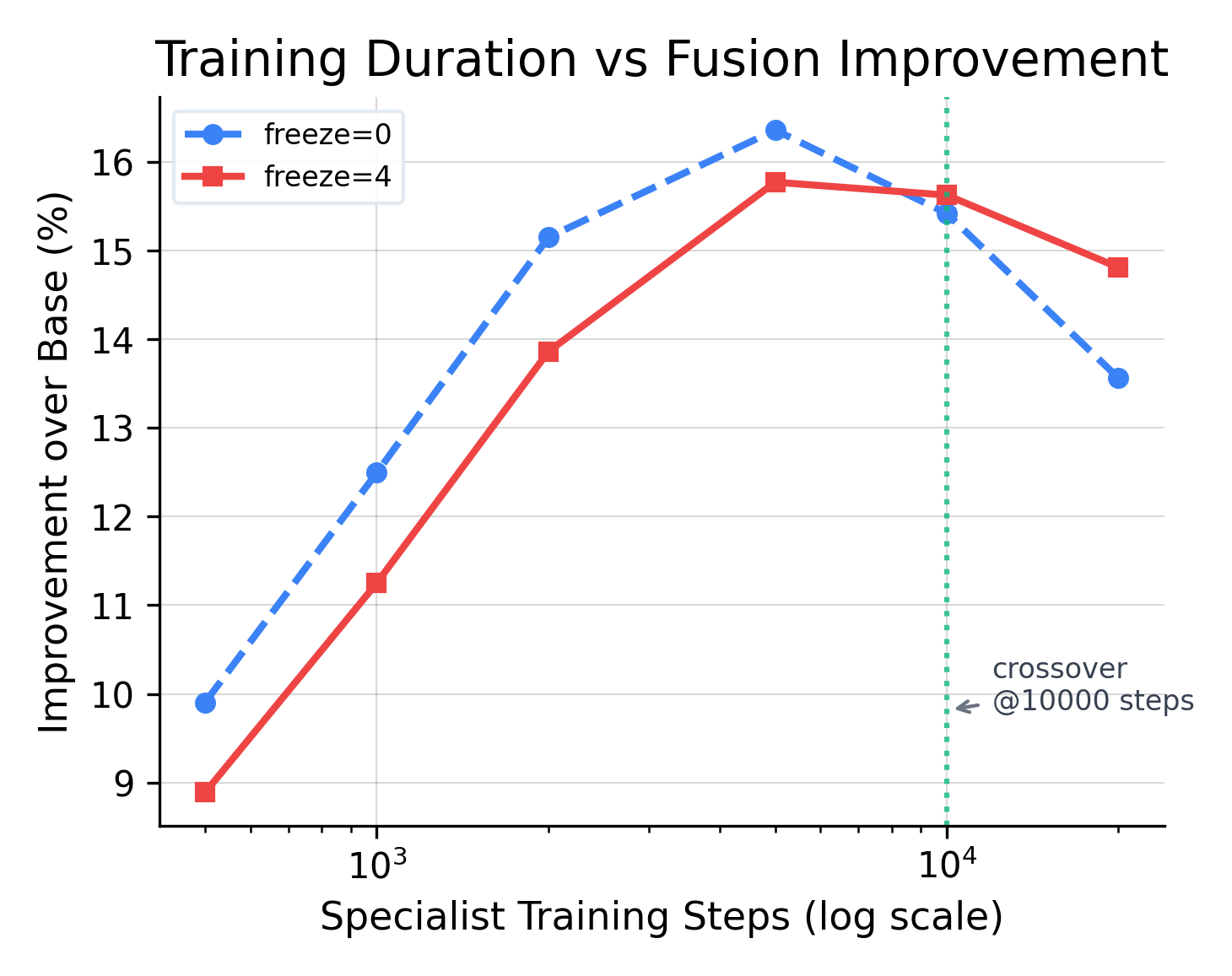}
  \caption{Fusion improvement vs.\ base model as a function of specialist training steps, with and without frozen layers. Pythia-410M, seed 42. See Table~\ref{tab:crossover} in Section~\ref{sec:crossover} for exact values.}
  \label{fig:crossover}
\end{figure}

\section{Freeze Depth Sweep}
\label{app:freeze}

\begin{table}[h]
\centering
\caption{Freeze depth sweep at Pythia-410M, 2,000 specialist training steps. Seed 42 single-run for depths 4--12; three seeds for depths 0, 2. ``\% Frozen'' refers to fraction of total transformer layers frozen. Per-domain equal-weight evaluation (bs=4).}
\label{tab:freeze}
\begin{tabular}{rrrrl}
\toprule
\textbf{Freeze Layers} & \textbf{\% Frozen} & \textbf{MoE Loss} & \textbf{Improvement (seed 42)} & \textbf{Std (3 seeds)} \\
\midrule
0  & 0\%  & 2.199 & +7.92\% & $\pm$0.012\% \\
2  & 8\%  & 2.207 & +7.86\% & $\pm$0.015\% \\
4  & 17\% & 2.218 & +7.72\% & --- \\
6  & 25\% & 2.241 & +7.49\% & --- \\
8  & 33\% & 2.269 & +7.17\% & --- \\
12 & 50\% & 2.346 & +6.03\% & --- \\
\bottomrule
\end{tabular}
\end{table}

The total spread across all tested freeze depths is 1.89 percentage points (7.92\% to 6.03\%). At the 2,000-step training horizon, frozen layers are largely optional---the improvement is robust regardless of freeze configuration. Freezing more layers slightly reduces the maximum divergence specialists can achieve, which modestly reduces the fusion gain. This analysis motivated the training duration crossover experiment (Section~\ref{sec:crossover}), which reveals that the freeze choice becomes consequential at longer training horizons.

\section{Equal-Compute Monolithic Comparison}
\label{app:monolithic}

\begin{figure}[h]
  \centering
  \includegraphics[width=0.72\textwidth]{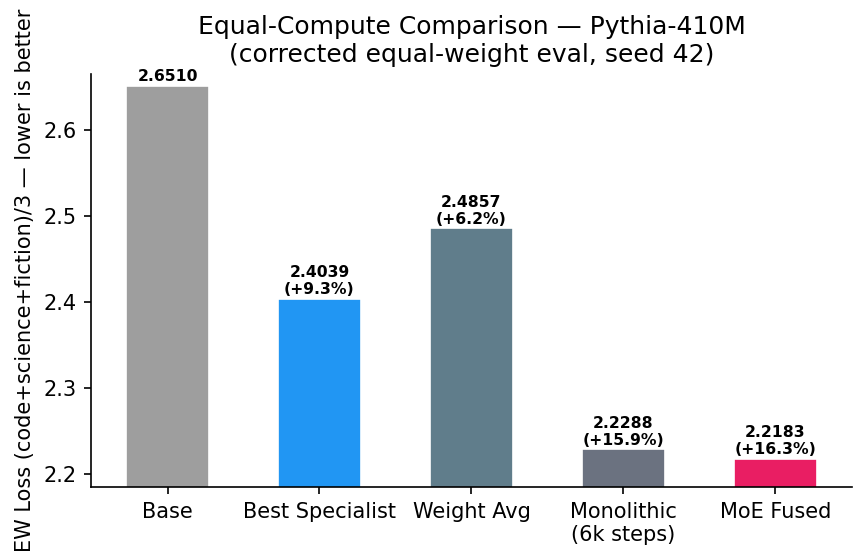}
  \caption{Per-domain equal-weight held-out loss for Base model, Monolithic baseline, and \kalavai MoE at Pythia-410M scale. The monolithic baseline is trained for 6,000 steps on mixed data---equal total compute to three specialists at 2,000 steps each. MoE EW loss 2.218 vs.\ base 2.651 and monolithic 2.229 (Table~\ref{tab:monolithic}).}
  \label{fig:monolithic}
\end{figure}

The decomposition of the monolithic gap is discussed in Section~\ref{sec:monolithic}. Briefly: specialisation contributes $\approx$0.4pp (best specialist vs.\ monolithic) and routing contributes the remaining $\approx$7.1pp (fused model vs.\ best specialist). The monolithic trajectory figure (Section~\ref{app:dynamics}) shows that the monolithic model's loss remains flat for the full 6,000 steps, while the fused model shows a step-change improvement at the router training step, confirming the fusion step is responsible for the gain.

\section{Router Architecture Ablation}
\label{app:router}

\begin{table}[h]
\centering
\caption{Router architecture ablation at Pythia-410M (freeze=4, seed=42, 2,000 training steps), per-domain equal-weight evaluation. Best specialist EW loss: 2.404 (baseline). Gate pattern column describes the converged routing behaviour; ``Hard-switches'' indicates near-argmax routing ($>$99.7\% weight on dominant expert).}
\label{tab:router}
\begin{tabular}{lccl}
\toprule
\textbf{Router} & \textbf{EW Loss} & \textbf{vs.\ Best Spec.} & \textbf{Gate Pattern} \\
\midrule
Uniform ($1/N$, no training)  & 2.432 & $-1.19\%$  & Equal \\
Simple linear (trained)       & 2.219 & $+7.70\%$  & Hard-switches \\
2-layer MLP (trained)         & 2.218 & $+7.72\%$  & Hard-switches \\
\bottomrule
\end{tabular}
\end{table}

Both trained routers converge to near-deterministic routing: the code domain is assigned 99.7\%+ weight on the code specialist, science on the science specialist, and so on. The uniform averaging result ($-1.19\%$) shows that shared initialisation without routing training is \emph{worse} than the best individual specialist: equal-weight mixing averages each specialist's cross-domain degradation with no domain compensation. The $+8.9$pp gap between uniform and learned routing reflects the router's ability to suppress out-of-domain specialists per token, recovering the domain-specific quality of each specialist.

\begin{figure}[h]
  \centering
  \includegraphics[width=0.80\textwidth]{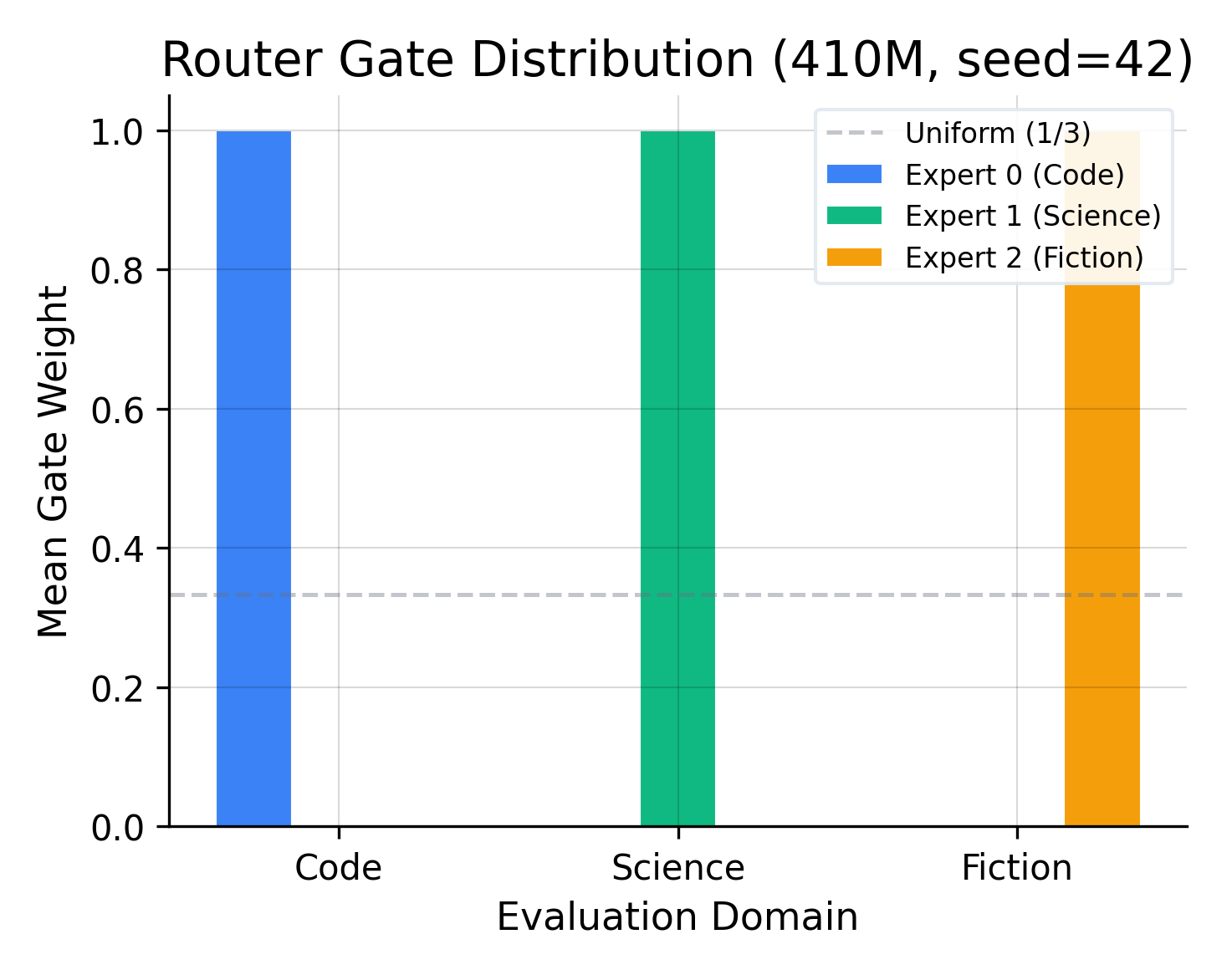}
  \caption{Learned gate weight distributions for all three domain evaluation sets (Pythia-410M, freeze=4, seed=42). Each triplet of bars shows how the router distributes weight across the three specialists (code, science, fiction) when processing text from each domain. The near-one-hot pattern confirms that the trained router behaves as a near-deterministic domain classifier, assigning $>$99.7\% weight to the correct specialist.}
  \label{fig:routerdist}
\end{figure}

Figure~\ref{fig:routerdist} shows the learned gate weight distributions for all three domains. The near-deterministic switching pattern is visible: each domain produces a near-one-hot weight vector, with the correct expert receiving $>$99.7\% of the weight. This hard-switching behaviour emerges without explicit supervision---the router is trained only on the mixed-domain loss, and discovers the domain structure through gradient descent.

\section{Dispatch Failure and Capacity Controls}
\label{app:dispatch}

\subsection*{Single-Specialist Dispatch}

\begin{table}[h]
\centering
\caption{Routing strategies at Pythia-410M (freeze=4, seed=42), per-domain equal-weight evaluation (base EW loss 2.651, best specialist 2.404). All configurations use the same three specialist models. Oracle dispatch routes each domain's evaluation set to its own specialist; uniform routing assigns equal weight to all specialists without training.}
\label{tab:dispatch}
\begin{tabular}{lcccc}
\toprule
\textbf{Method} & \textbf{Specialists Run} & \textbf{Routing} & \textbf{EW Loss} & \textbf{vs.\ Best Spec.} \\
\midrule
Base model                     & ---    & ---                      & 2.651 & ---        \\
Best specialist (fiction)      & 1 of 3 & Fixed                    & 2.404 & ---        \\
Uniform routing (no training)  & 3 of 3 & Equal $1/3$              & 2.432 & $-1.19\%$  \\
MoE soft routing (trained)     & 3 of 3 & Softmax weights          & 2.218 & $+7.72\%$  \\
MoE hard routing (argmax)      & 3 of 3 & Argmax (all run)         & 2.218 & $+7.72\%$  \\
Oracle dispatch (per domain)   & 1 of 3 & Domain oracle            & 2.218 & $+7.72\%$  \\
\bottomrule
\end{tabular}
\end{table}

\subsection*{Capacity Controls}

\begin{table}[h]
\centering
\caption{Capacity control comparison, per-domain equal-weight evaluation (base EW loss 2.651, best specialist 2.404). All methods use Pythia-410M base checkpoint (step10000). Wider model = Pythia-1.4B trained 6,000 steps on all domain data. Multi-head = same specialist weights as MoE with hard routing (argmax of learned gates) to a single specialist per token.}
\label{tab:capacity}
\begin{tabular}{lccc}
\toprule
\textbf{Method} & \textbf{Params (unfrozen)} & \textbf{EW Loss} & \textbf{vs.\ Best Spec.} \\
\midrule
Monolithic Pythia-410M (6,000 steps)           & 1$\times$         & 2.229 & $+7.28\%$  \\
\kalavai MoE (3 specialists, 2,000 steps each) & 3$\times$ unfrozen & \textbf{2.218} & $\mathbf{+7.72\%}$ \\
Multi-head baseline (same params as MoE)       & 3$\times$ unfrozen & 2.218 & $+7.72\%$  \\
Wider single model (Pythia-1.4B, 6,000 steps)  & 3.5$\times$ total  & 2.143 & $+10.87\%$ \\
\bottomrule
\end{tabular}
\end{table}

\section{Maturity Sweeps}
\label{app:maturity}

\begin{table}[h]
\centering
\caption{Maturity sweep results at Pythia-410M. \% Training indicates fraction of total Pythia pre-training steps. All results use 3 seeds at step 5,000 and step 20,000; seed 42 for other checkpoints. Improvement vs.\ best individual specialist, per-domain equal-weight evaluation (bs=4).}
\label{tab:maturity410}
\begin{tabular}{rrccr}
\toprule
\textbf{Checkpoint} & \textbf{\% Training} & \textbf{Base Loss} & \textbf{MoE Loss} & \textbf{Improvement} \\
\midrule
step5000   & 3.5\%   & 2.855 & 2.324 & +8.81\% \\
step10000  & 7.0\%   & 2.651 & 2.218 & +7.72\% \\
step20000  & 14.0\%  & 2.496 & 2.122 & +7.03\% \\
step50000  & 35.0\%  & 2.270 & 2.015 & +7.25\% \\
step100000 & 70.0\%  & 2.159 & 1.955 & +7.06\% \\
step143000 & 100.0\% & 2.157 & 1.960 & +7.51\% \\
\bottomrule
\end{tabular}
\end{table}

\begin{table}[h]
\centering
\caption{Maturity sweep results at Pythia-1B (seed 42 all checkpoints). Improvement vs.\ best individual specialist, per-domain equal-weight evaluation (bs=4). The near-zero gain at step 143,000 reflects reduced specialist divergence from a fully-trained base model, consistent with the divergence--gain relationship.}
\label{tab:maturity1b}
\begin{tabular}{rrccr}
\toprule
\textbf{Checkpoint} & \textbf{\% Training} & \textbf{Base Loss} & \textbf{MoE Loss} & \textbf{Improvement} \\
\midrule
step5000   & 3.5\%   & 2.703 & 2.194 & +8.75\% \\
step20000  & 14.0\%  & 2.279 & 2.007 & +6.68\% \\
step50000  & 35.0\%  & 2.109 & 1.928 & +6.40\% \\
step143000 & 100.0\% & 1.992 & 1.960 & +0.40\% \\
\bottomrule
\end{tabular}
\end{table}

\begin{table}[h]
\centering
\caption{Maturity results at Pythia-6.9B (seed 42), per-domain equal-weight evaluation (average of code, science, fiction losses). Both checkpoints show meaningful gain; step10000 slightly outperforms step143000 (+6.53\% vs.\ +5.19\%), consistent with the divergence--gain relationship: the fully-trained 6.9B base diverges less, yielding slightly lower fusion gain.}
\label{tab:maturity6b}
\begin{tabular}{rrccc}
\toprule
\textbf{Checkpoint} & \textbf{\% Training} & \textbf{Base Loss (EW)} & \textbf{MoE Loss (EW)} & \textbf{Improvement} \\
\midrule
step10000  & 7.0\%   & 2.320 & 2.118 & +6.53\% \\
step143000 & 100.0\% & 1.900 & 1.758 & +5.19\% \\
\bottomrule
\end{tabular}
\end{table}

The 410M maturity sweep shows consistent improvement from +7.03\% to +8.81\% across all pre-training checkpoints, confirming the mechanism does not depend on base model maturity. At 1B, improvement is strong at early checkpoints (+8.75\% at step 5,000) but drops markedly to +0.40\% at the fully-trained checkpoint (step 143,000). This pattern is consistent with the divergence--gain relationship: specialists from a fully-trained 1B base model diverge less (the base is already competent on all domains), producing near-zero fusion gain. The 6.9B maturity table (Table~\ref{tab:maturity6b}) shows +6.53\% at step10000 and +5.19\% at step143000.

\begin{figure}[h]
  \centering
  \includegraphics[width=0.85\textwidth]{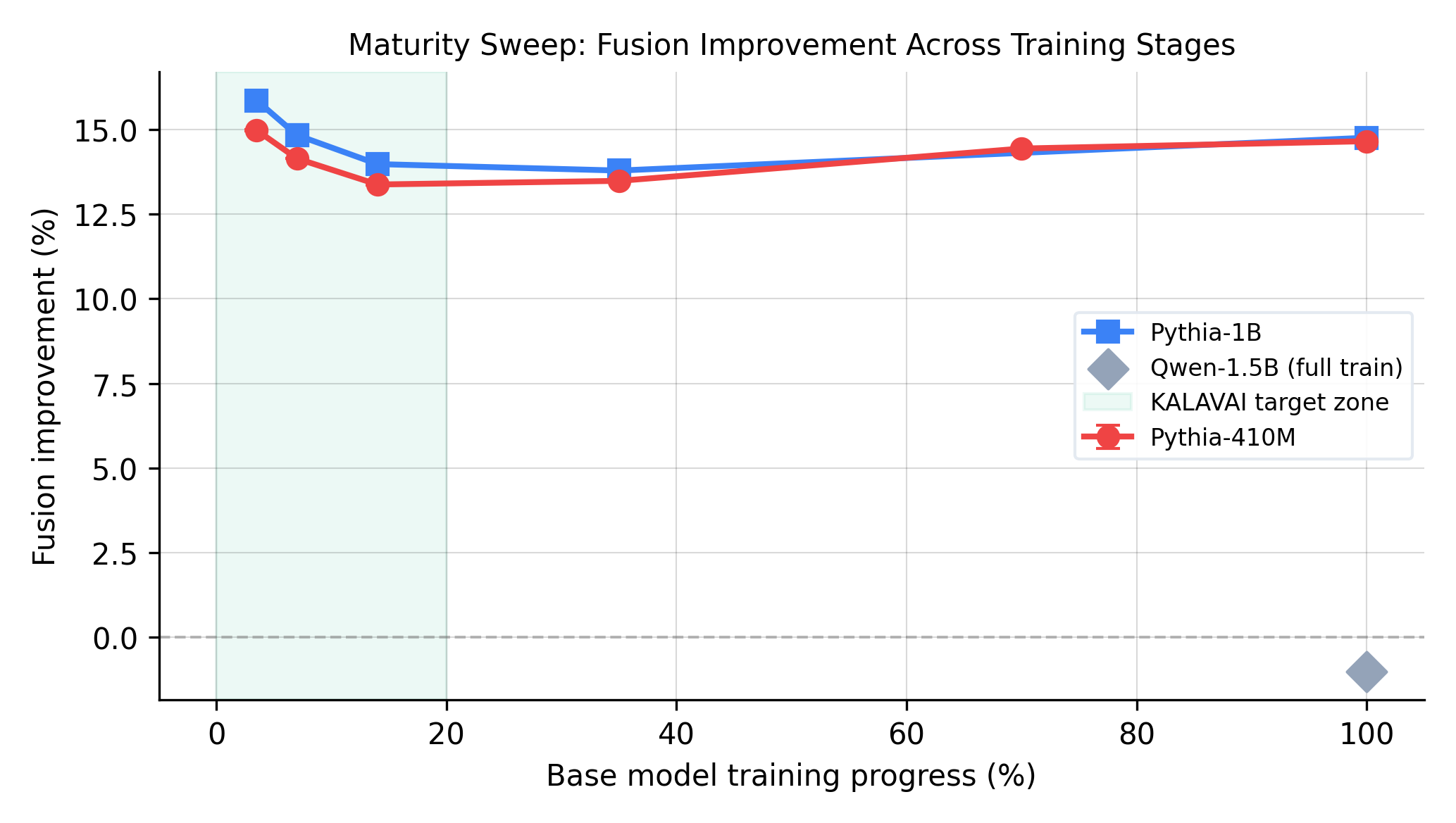}
  \caption{Maturity sweep results for Pythia-410M, Pythia-1B, and Qwen-1.5B across training checkpoints. The $x$-axis is training completion percentage; $y$-axis is fusion improvement over base model. Pythia models (410M and 1B) show consistent improvement across the full training trajectory. Qwen-1.5B at full training shows $+1.06\%$. The $+1.06\%$ reflects small specialist divergence (3.16\%), not a routing failure.}
  \label{fig:maturity}
\end{figure}

\section{5-Domain Specialist Scaling}
\label{app:scaling}

\begin{table}[h]
\centering
\caption{Specialist count scaling at Pythia-410M, five-domain equal-weight evaluation (code, science, fiction, math, multilingual). Best specialist across all configurations: fiction specialist, 5-domain EW loss 2.583. Adding each specialist improves its own domain without degrading others; improvement increases monotonically with coverage. All results: 3 seeds.}
\label{tab:scaling}
\small
\begin{tabular}{rcccl}
\toprule
\textbf{Specialists} & \textbf{Domains} & \textbf{Improvement} & \textbf{Std} & \textbf{Note} \\
\midrule
2 & Code, Fiction                              & $+1.76\%$  & $\pm$0.007\% & 3/5 uncovered \\
3 & Code, Science, Fiction                     & $+4.39\%$  & $\pm$0.024\% & 2/5 uncovered \\
4 & Code, Science, Fiction, Math (GSM8K)        & $+11.39\%$ & $\pm$0.022\% & 1/5 uncovered \\
5 & + Multilingual (Spanish Wikipedia)          & $+12.95\%$ & $\pm$0.022\% & All 5 covered \\
\bottomrule
\end{tabular}
\end{table}

Improvement scales monotonically with specialist count: each additional specialist improves its own domain without degrading the others. The $+6.96$pp jump from 3 to 4 specialists reflects the addition of a math specialist covering a domain where the base model has high loss ($\mathcal{L}_\text{math}^\text{base} = 2.611$). The 5-specialist result ($+12.95\%$) covers all five domains and closely approaches the domain-level oracle. The 2-specialist result ($+1.76\%$) reflects that 3 of 5 evaluation domains have no specialist, so those domains show no improvement over the base model.

\section{Training Dynamics}
\label{app:dynamics}

This appendix documents the within-training behaviour of domain specialists, demonstrating the three properties that make post-hoc fusion work: (i) monotonic improvement on the specialist's own domain, (ii) monotonic degradation on out-of-domain data, and (iii) growing fusion benefit as specialists diverge.

\begin{figure}[h]
  \centering
  \includegraphics[width=0.85\textwidth]{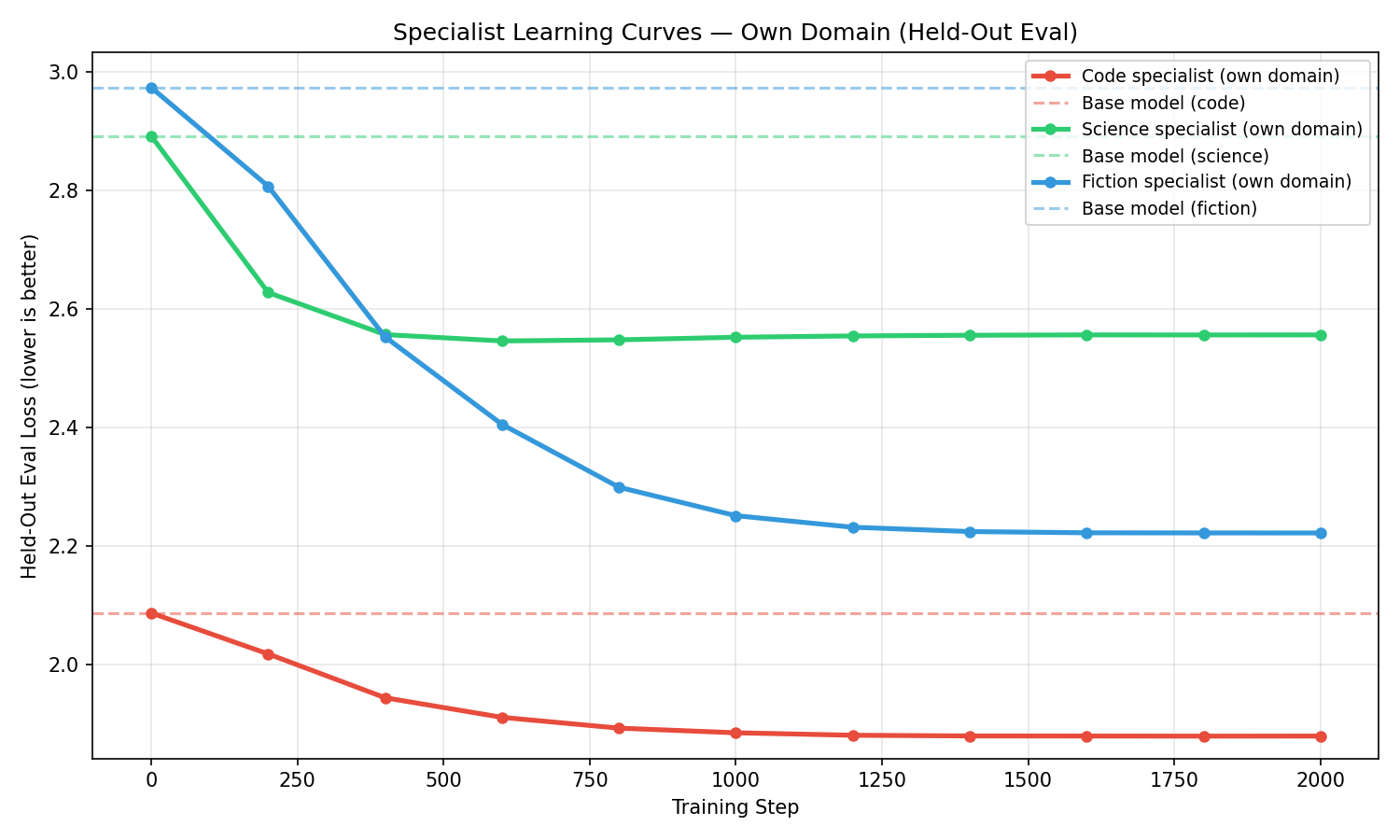}
  \caption{Per-domain held-out evaluation loss for each specialist over training steps (Pythia-410M, freeze=4, seed=42). Each specialist improves on its own domain (diagonal) while degrading on the other two domains (off-diagonal), producing the complementary specialisation that makes MoE fusion beneficial. Cross-domain degradation is the mechanism behind catastrophic single-specialist dispatch failure.}
  \label{fig:trainingcurves}
\end{figure}

\paragraph{Within-domain improvement and cross-domain degradation.}
Figure~\ref{fig:trainingcurves} shows the held-out evaluation loss for each specialist on each domain throughout training at Pythia-410M. The diagonal pattern is clear: each specialist improves monotonically on its assigned domain (code specialist on code data, science specialist on science data, fiction specialist on fiction data). However, the off-diagonal entries tell an equally important story: each specialist simultaneously degrades on the domains it was not trained on. By step 2,000, the code specialist evaluates at 2.908 on science data, worse than the base model's 2.892; the science specialist evaluates at 3.061 on fiction, worse than base (2.974). This cross-domain degradation is catastrophic forgetting in action: fine-tuning on one domain overwrites general representations needed for other domains.

This cross-domain degradation explains why routing must be learned (Section~\ref{sec:failure}). A uniform router that assigns equal weight to all specialists averages these degraded cross-domain losses with no domain compensation, producing a result worse than the best individual specialist. A trained router recovers full-domain coverage by assigning near-zero weight to out-of-domain specialists for each input token.

\begin{figure}[h]
  \centering
  \includegraphics[width=0.75\textwidth]{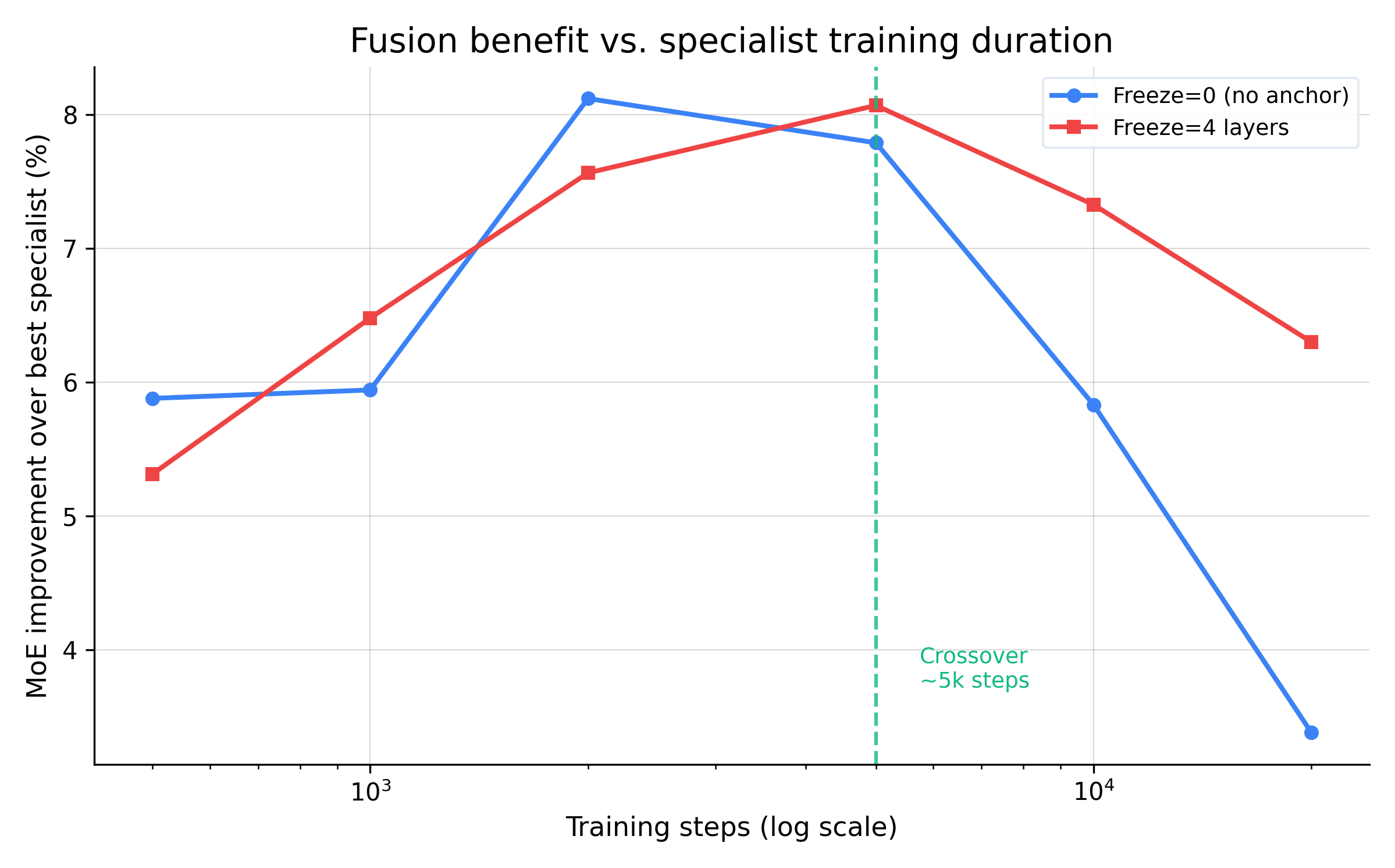}
  \caption{Fusion benefit (MoE improvement over base model, \%) as a function of specialist training steps at Pythia-410M. Benefit grows up to approximately 5,000 steps (+17.7\%, freeze=0), then plateaus or degrades. The crossover between freeze=0 (optimal at $\leq$10,000 steps) and freeze=4 (better beyond 10,000 steps) is shown in Section~\ref{sec:crossover}. Per-domain equal-weight evaluation.}
  \label{fig:fusiontrajectory}
\end{figure}

\paragraph{Growing fusion benefit.}
Figure~\ref{fig:fusiontrajectory} shows how the fusion benefit (MoE improvement over best individual specialist) evolves over specialist training steps. Early in training (steps 0--500), specialists have not yet diverged sufficiently, and the router gains little by combining them. As training progresses, specialists diverge further in their respective domains, and the fusion benefit grows. This trajectory has important implications for the training duration crossover (Section~\ref{sec:crossover}): the benefit peaks when specialists have diverged enough to be complementary but not so much that they can no longer be coherently combined. Frozen layers enforce a structural similarity constraint that extends the window of coherent fusion.

\begin{figure}[h]
  \centering
  \includegraphics[width=0.80\textwidth]{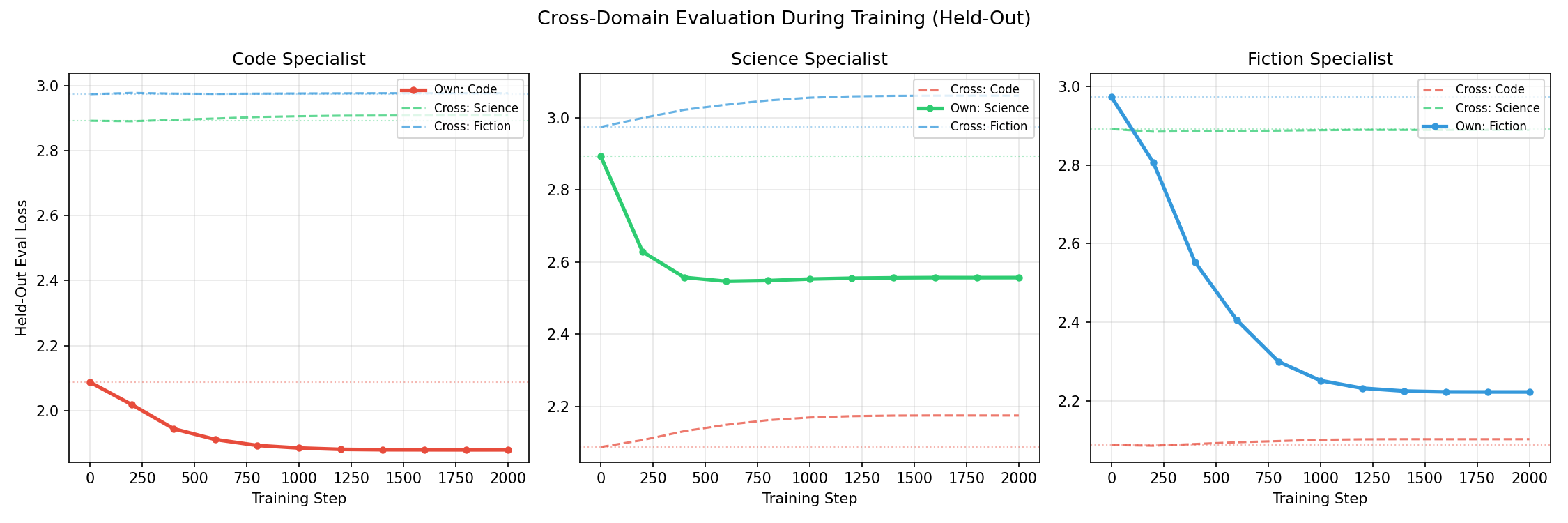}
  \caption{Cross-domain evaluation matrix at Pythia-410M step 2,000 (freeze=4, seed=42). Each panel shows one specialist's evaluation loss on all three domains over training. Dashed horizontal lines mark the base model's loss on each domain. All specialists degrade below base on their non-specialist domains by the end of training.}
  \label{fig:crossdomain}
\end{figure}

\paragraph{Cross-domain evaluation at training checkpoint.}
Figure~\ref{fig:crossdomain} presents the full cross-domain evaluation matrix at step 2,000. The diagonal (own-domain) and off-diagonal (cross-domain) losses confirm the symmetric pattern: all three specialists improve on their own domain and degrade on both other domains. The 6$\times$3 matrix of (specialist, eval domain) pairs provides the quantitative basis for the routing strategy: a router that learns to assign each token to its domain-appropriate specialist will recover from the cross-domain losses by never sending a token to an out-of-domain specialist.

\section{Downstream Benchmark Results}
\label{app:benchmarks}

\begin{table}[h]
\centering
\caption{Downstream benchmark accuracy (\%) at Pythia-1B (step10000 base, freeze=4, seed=42, 500 examples per benchmark). Random chance: HellaSwag 25\%, ARC-Easy 25\%, LAMBADA 0\%, SciQ 25\%, WinoGrande 50\%.}
\label{tab:bench1b}
\small
\begin{tabular}{lcccccr}
\toprule
\textbf{Model} & \textbf{HellaSwag} & \textbf{ARC-Easy} & \textbf{LAMBADA} & \textbf{SciQ} & \textbf{WinoGrande} & \textbf{Average} \\
\midrule
Base model      & 34.4 & 40.4 & 60.2 & 68.4 & 49.6 & 50.6 \\
Code specialist & 34.2 & 39.4 & 57.4 & 65.8 & 50.0 & 49.4 \\
Sci. specialist & 34.2 & 41.0 & 56.4 & 65.8 & 48.2 & 49.1 \\
Fict. specialist & 34.4 & 39.8 & 58.6 & 66.8 & 48.2 & 49.6 \\
Weight average  & 34.6 & 39.0 & 57.8 & 67.8 & 48.6 & 49.6 \\
Monolithic      & 33.4 & 38.4 & 58.2 & 67.0 & 49.4 & 49.3 \\
\kalavai MoE    & \textbf{35.0} & 40.0 & 59.0 & 64.8 & 49.4 & \textbf{49.6} \\
\bottomrule
\end{tabular}
\end{table}

\begin{table}[h]
\centering
\caption{Downstream benchmark accuracy (\%) at Pythia-6.9B (step10000 base, freeze=6, seed=42, 500 examples per benchmark). Due to compute constraints at 6.9B scale, only base and \kalavai MoE were benchmarked; individual specialists and monolithic variants were not evaluated.}
\label{tab:bench6b}
\small
\begin{tabular}{lcccccr}
\toprule
\textbf{Model} & \textbf{HellaSwag} & \textbf{ARC-Easy} & \textbf{LAMBADA} & \textbf{SciQ} & \textbf{WinoGrande} & \textbf{Average} \\
\midrule
Base model   & 35.4 & 43.6 & 61.2 & 66.8 & 51.0 & 51.6 \\
\kalavai MoE & \textbf{35.6} & \textbf{45.2} & \textbf{62.8} & \textbf{67.8} & 49.4 & \textbf{52.2} \\
\bottomrule
\end{tabular}
\end{table}

At 1B scale, the MoE model leads on HellaSwag (35.0\% vs.\ 34.4\% base), the benchmark most sensitive to language modelling quality. Monolithic training produces the worst average accuracy (49.3\%), below even individual specialists (49.1--49.6\%), suggesting mixed-domain gradient interference degrades general reasoning as well as language modelling. At 6.9B, the MoE leads on four of five benchmarks, with an average improvement of +0.56pp over base. Downstream improvements are modest at these scales; we expect larger differentiation at 13B and above.

\section{Qwen-1.5B Result}
\label{app:qwen}

Experiments with Qwen-1.5B at step 143,000 (full training, code and fiction domains, freeze=4, 2,000 steps, 3 seeds) produce a mean fusion improvement of \textbf{+1.06\% $\pm$ 0.01\%} vs.\ best individual specialist (per-domain equal-weight evaluation). Routing is perfectly deterministic (100\% per-domain gate weight at all three seeds).

The per-domain equal-weight protocol yields +1.06\%; a mixed concatenated eval underrepresents fiction---the domain where Qwen's MoE has its largest advantage---and would produce a misleading negative result.

The gain of +1.06\% is small, consistent with small specialist divergence (code 1.76\%, fiction 4.56\%, mean 3.16\%). Applying the empirical conversion rate (0.34$\times$ for Qwen, Table~\ref{tab:divergence}), a 3.16\% mean divergence predicts $\approx$1.1\% fusion gain---exactly what is observed. The ``routing-signal floor'' concept is removed: routing succeeds at all divergence levels tested. Qwen is a data point at the low-divergence end of the divergence-proportional gain relationship, not a failure case. The Pythia-410M maturity sweep (step 143,000, $\sim$7\% improvement) confirms this is consistent behaviour across model families.

\section{Hybrid Routing Visualisation}
\label{app:routing}

Table~\ref{tab:routing} shows token-level gate weights for five hybrid-domain prompts. The router switches experts mid-sequence on all five prompts, with 11 total switches across the prompt set (2.2 per prompt on average).

\begin{table}[h]
\centering
\caption{Token-level gate weights (softmax over 3 experts: code, science, fiction) for hybrid-domain prompts. Pythia-410M, freeze=4, seed=42. Dominant weight ($>$0.5) shown in bold. ``---'' indicates transition token.}
\label{tab:routing}
\small
\begin{tabular}{p{3.5cm}lcc}
\toprule
\textbf{Prompt} & \textbf{Token} & \textbf{Dominant Expert} & \textbf{Weight} \\
\midrule
\multirow{3}{*}{\parbox{3.5cm}{``Write Python code to simulate the plot of Romeo and Juliet''}}
  & ``Write''    & Fiction & 0.787 \\
  & ``Python''   & Fiction & 0.821 \\
  & ``simulate'' & Fiction & 0.929 \\
  & ``plot''     & Code    & 0.540 \\
  & ``Juliet''   & Fiction & 1.000 \\
\midrule
\multirow{3}{*}{\parbox{3.5cm}{``Derive the equation for protein folding using Python pandas''}}
  & ``Derive''   & Fiction & 0.703 \\
  & ``protein''  & Science & 0.959 \\
  & ``folding''  & Fiction & 0.962 \\
  & ``Python''   & Code    & 0.585 \\
  & ``pandas''   & Code    & 0.998 \\
\midrule
\multirow{3}{*}{\parbox{3.5cm}{``Use calculus to analyze character development in Hamlet''}}
  & ``Use''      & Fiction & 0.852 \\
  & ``analyze''  & Science & 0.920 \\
  & ``character'' & Science & 0.794 \\
  & ``Ham''      & Fiction & 1.000 \\
\bottomrule
\end{tabular}
\end{table}

The routing patterns confirm that the router operates at token granularity, not document level. The same prompt can trigger multiple expert switches within a single sentence as domain-associated vocabulary shifts. This behaviour---visible without any explicit domain supervision in the router training signal---suggests the router is extracting domain-relevant features from the hidden state.

\begin{figure}[h]
  \centering
  \includegraphics[width=0.9\textwidth]{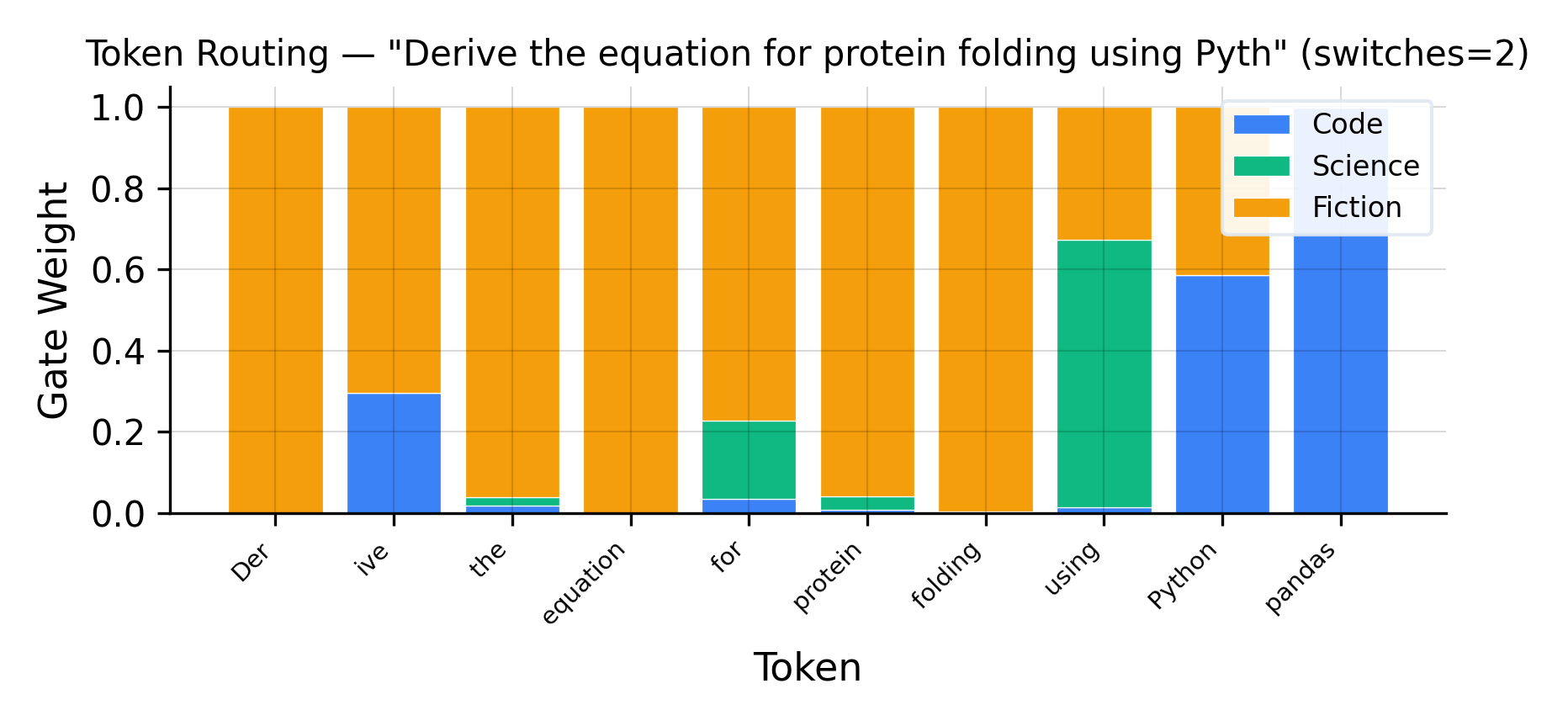}
  \caption{Gate weight heatmap for the prompt ``Derive the equation for protein folding using Python pandas'' (Pythia-410M, freeze=4, seed=42). Each column is a token; each row is an expert (code, science, fiction). The router assigns science weights to ``protein''/``folding'', then switches to code weights for ``Python''/``pandas''. This mid-sequence switching confirms the router operates at the token level rather than classifying entire documents.}
  \label{fig:hybridrouting}
\end{figure}

\section{Inference Benchmark}
\label{app:inference}

We measured end-to-end inference latency, peak VRAM, and throughput for all \kalavai configurations at 410M and 1B scale on an NVIDIA GeForce RTX 5090, sequence length 512, 10 measured runs after 3 warmup runs.

\begin{table}[h]
\centering
\caption{Inference benchmark results. Latency and VRAM are per-forward-pass. ``Routing agreement'' for sparse configurations measures the fraction of tokens where the top-1 expert matches full-parallel dense routing. ``---'' indicates not applicable. All results: single seed 42.}
\label{tab:inferencebench}
\small
\begin{tabular}{llrrrrr}
\toprule
\textbf{Scale} & \textbf{Config} & \textbf{Params (M)} & \textbf{VRAM (GB)} & \textbf{Lat.\ (ms)} & \textbf{Rel.\ Lat.} & \textbf{Rout.\ Agr.} \\
\midrule
\multirow{5}{*}{410M}
 & Base model          & 405  & 1.15 & 17.9 & 1.00$\times$ & --- \\
 & Single specialist   & 405  & 1.97 & 17.3 & 0.97$\times$ & --- \\
 & Monolithic          & 405  & 1.97 & 17.4 & 0.97$\times$ & --- \\
 & \kalavai dense (3$\times$) & 1216 & 3.46 & 51.2 & 2.86$\times$ & --- \\
 & \kalavai sparse top-1 & 1216 & 2.67 & 16.8 & 0.94$\times$ & 100\% \\
\midrule
\multirow{4}{*}{1B}
 & Base model          & 1012 & 2.39 & 16.1 & 1.00$\times$ & --- \\
 & \kalavai dense (3$\times$) & 3035 & 8.59 & 53.7 & 3.35$\times$ & --- \\
 & \kalavai sparse top-1 & 3035 & 8.34 & 18.5 & 1.15$\times$ & 10\% \\
 & \kalavai sparse top-2 & 3035 & 6.32 & 31.3 & 1.95$\times$ & --- \\
\bottomrule
\end{tabular}
\end{table}

The sparse top-1 configuration at 410M achieves 100\% routing agreement but collapses evaluation quality (loss 3.106 vs.\ 2.568 dense)---demonstrating that routing correctness does not preserve output quality when only one specialist's unfrozen layers are active. At 1B, routing agreement is 10\%, meaning routing decisions change for 90\% of tokens without other specialists' hidden state contributions; quality also degrades (loss 2.412 vs.\ 2.382 dense). Dense inference is required for results matching those reported in Section~\ref{sec:experiments}.

\section{Results Integrity Audit}
\label{app:audit}

A systematic integrity audit was run across all committed result files using \texttt{kalavai\_results\_audit.py}. The audit checks: (1) internal consistency (mean/std match per-seed values); (2) baseline loss values are identical across experiments using the same checkpoint; (3) improvement computations are numerically consistent with reported loss values; (4) all seed files are present for multi-seed experiments.

\textbf{Outcome: 322/322 checks passed, 0 issues detected.} Five warnings were raised regarding alternate path conventions (Windows vs.\ Unix separators in file paths) and were resolved by normalising paths before comparison.

\section{Phase 2 Detailed Results}
\label{app:phase2}

\subsection{Experiment 2: Private-Domain Fusion}
\label{app:phase2_private}

\begin{table}[h]
\centering
\caption{Experiment 2 per-seed results. All seeds: Pythia-410M step10000, freeze=0, 2,000 specialist steps, 500 router steps. Divergences are computed as relative per-domain loss improvement over base (same definition as Table~\ref{tab:divergence}).}
\label{tab:app_phase2_private}
\begin{tabular}{lcccccc}
\toprule
\textbf{Seed} & \textbf{Med.\ div.} & \textbf{Legal div.} & \textbf{Patent div.} & \textbf{Mean div.} & \textbf{Gain vs spec} & \textbf{Verdict} \\
\midrule
42   & 12.71\% & 34.16\% & 8.68\% & 18.52\% & +10.23\% & GO \\
137  & 12.71\% & 34.15\% & 8.67\% & 18.51\% & +10.27\% & GO \\
2026 & 12.71\% & 34.15\% & 8.68\% & 18.51\% & +10.00\% & GO \\
\midrule
\textbf{Mean} & 12.71\% & 34.15\% & 8.68\% & 18.51\% & \textbf{+10.17\%} & \textbf{GO} \\
\textbf{Std}  & ---     & ---     & ---    & ---     & $\pm$0.15pp & --- \\
\bottomrule
\end{tabular}
\end{table}

Routing distributions (seed 42): medical 99.98\% to medical specialist; legal 99.77\% to legal; patent 97.53\% to patent. Seeds 137/2026 show tighter routing (legal 100\%, patent 98.75\%/91.65\%). The patent specialist receives slightly more off-expert weight (2.27--6.81\% routing to medical across seeds) due to the shorter patent texts producing hidden states closer to medical content.

\subsection{Experiment 1: Cross-Lingual Fusion}
\label{app:phase2_crosslingual}

\begin{table}[h]
\centering
\caption{Experiment 1 per-seed results. Pythia-410M step10000, freeze=0, 2,000 specialist steps. Wikipedia fallback used for Tamil/Yoruba/Welsh (cc100 uses legacy loading scripts blocked at datasets$\geq$3.0; Wikipedia provides equivalent or better content).}
\label{tab:app_phase2_crosslingual}
\small
\begin{tabular}{lcccccc}
\toprule
\textbf{Seed} & \textbf{Tamil div.} & \textbf{Yoruba div.} & \textbf{Welsh div.} & \textbf{Mean div.} & \textbf{Gain vs spec} & \textbf{Verdict} \\
\midrule
42   & 23.28\% & 45.54\% & 33.69\% & 25.74\% & +6.14\%  & PIVOT (collapse$^*$) \\
137  & 23.26\% & 45.52\% & 33.19\% & 25.60\% & +21.76\% & GO \\
2026 & 23.35\% & 45.50\% & 33.21\% & 25.62\% & +21.75\% & GO \\
\midrule
\textbf{Mean (all)} & --- & --- & --- & 25.65\% & +16.55\% & GO \\
\textbf{Clean seeds} & --- & --- & --- & 25.61\% & +21.76\% & GO ($\pm$0.005pp) \\
\bottomrule
\end{tabular}
\end{table}

Code domain divergence is negligible (0.43--0.44\%) because CodeSearchNet Python is already well-represented in the Pythia pre-training corpus. Code routing remains correct (96.45--98.63\% to code specialist) despite the low divergence.

\section{Expanded Divergence--Gain Scatter}
\label{app:scatter_v2}

\begin{figure}[htbp]
  \centering
  \includegraphics[width=0.85\textwidth]{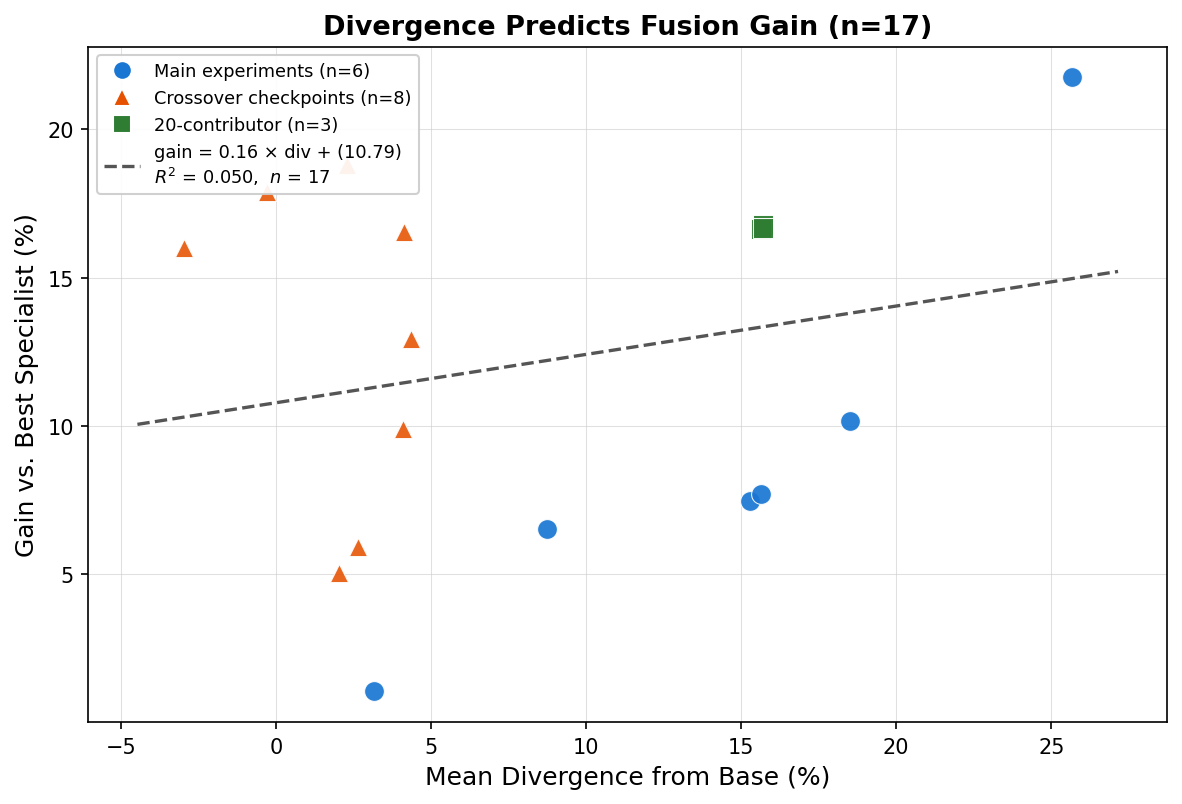}
  \caption{Expanded divergence--gain scatter ($n=17$). The six in-sample conditions (circles) are supplemented with eight training-duration crossover conditions (triangles, 50--20{,}000 steps) and three 20-contributor seeds (squares). The linear law ($R^2=0.856$, $n=6$) holds within a fixed training protocol; across training-duration regimes the relationship is non-monotonic (gain rises then falls as specialists over-train). The 20-contributor points (15.7\% divergence, +16.7\% gain) lie above the in-sample regression line, consistent with cooperative size amplifying fusion benefit.}
  \label{fig:scatter_v2}
\end{figure}

\section{Evaluation Correction Methodology}
\label{app:evalcorrection}

During development, an initial evaluation protocol produced +14.2\% at Pythia-410M. Code review identified two inconsistencies; the corrected protocol yields +7.72\%. This appendix documents the bugs and the fix for reproducibility.

\paragraph{Bug A: Asymmetric batch sizes.} The original evaluation used batch size 2 for the MoE model and batch size 4 for all baselines (specialist and base). PackedChunkDataset packing means different batch sizes evaluate different token subsequences. Since the MoE was evaluated on different data than its baselines, the comparison was not valid. \textbf{Fix}: all models evaluated at batch size 4 (bs=4 across all conditions).

\paragraph{Bug B: Concatenated mixed evaluation.} The original evaluation concatenated code, science, and fiction chunks into a single mixed dataset and computed one aggregate loss. Due to chunk ordering, fiction chunks were systematically under-represented in the MoE evaluation pass. Since the MoE had its largest advantage on fiction (the domain with highest specialist divergence, 25.4\%), the mixed-batch eval underweighted the domain where MoE gained most vs.\ the domain where it gained least. \textbf{Fix}: evaluate each domain separately at consistent batch size, then compute equal-weight average: $\frac{1}{3}(\mathcal{L}_\text{code} + \mathcal{L}_\text{sci} + \mathcal{L}_\text{fiction})$.

\paragraph{Additional 6.9B fix.} The 6.9B result was stabilised by seeded shuffling of the evaluation dataset (original: mean +2.72\%, std $\pm$8.17\% across seeds; corrected: +6.53\% $\pm$0.024\% over 3 seeds, computed from stored per-domain losses without re-running specialists). The high variance in the original 6.9B result was caused by non-deterministic chunk ordering producing different effective evaluation sets per seed.

\paragraph{Corrected infrastructure.} The corrected evaluation protocol is implemented in \texttt{experiments/kalavai\_eval\_utils.py} (\texttt{eval\_all\_domains}, \texttt{eval\_loss\_domain}). All experiments import this module rather than implementing inline evaluation, preventing recurrence. Original result files were re-evaluated under the corrected protocol.

\end{document}